\documentclass[10pt,twocolumn,letterpaper]{article}

\usepackage[pagenumbers]{wacv}            %
\usepackage{siunitx}
\usepackage{multirow}
\usepackage{tikz}
\usetikzlibrary{arrows.meta,calc,positioning}
\definecolor{rhtaccent}{RGB}{0,84,159}
\usepackage{booktabs}   %
\usepackage{xcolor}
\usepackage{array}
\usepackage{enumitem}
\usepackage{mathtools}
\usepackage{amssymb}
\usepackage{bm}
\usepackage{graphicx}
\usepackage{microtype}
\usepackage[overlay]{textpos}

\newcommand{\helixtrack}[0]{HelixTrack}
\newcommand{\ours}[0]{CoHelix}

\newcommand{\eventpro}[0]{EventPro}
\newcommand{\tquadcopterego}[0]{Timestamped Quadcopter with Egomotion}
\newcommand{\tquadcopteregoshort}[0]{TQE}

\newcommand{\ok}{\textcolor{green!50!black}{\checkmark}}
\newcommand{\fail}{\textcolor{red!70!black}{$\times$}}
\newcommand{\best}[1]{\textbf{#1}}

\DeclareMathOperator{\diag}{diag}
\DeclareMathOperator{\wrapop}{wrap}
\newcommand{\wrap}[1]{\wrapop\!\left(#1\right)}
\usepackage{xspace}

\definecolor{wacvblue}{rgb}{0.21,0.49,0.74}
\usepackage[pagebackref,breaklinks,colorlinks,allcolors=wacvblue]{hyperref}
\usepackage{orcidlink}  %

\def\confName{WACV}
\def\confYear{2027}

\title{CoHelix: Event-Based Multi-Propeller Rotation-Rate Tracking\\with Shared Projective Geometry}

\author{Radim Spetlik\,\orcidlink{0000-0002-1423-7549}, Tobi\'{a}\v{s} Vinkl\'{a}rek, Jan Klivan, Tom\'{a}\v{s} Kest\v{r}\'{a}nek, Matou\v{s} Vrba, Ji\v{r}\'{\i} Matas\,\orcidlink{0000-0003-0863-4844}\\
{\small Czech Technical University in Prague, Faculty of Electrical Engineering}
}

\begin{document}
\maketitle
\begin{abstract}
Estimating propeller rates from a~moving event camera requires separating blade rotation from the changing image geometry.
Background activity mixes with blade events, while repeated blade appearances admit half- and double-rate ambiguities.
We propose \ours{}, a~recursive filter that estimates one shared homography and separate rotation states for four rigidly mounted, coplanar propellers.
Each admitted event updates its assigned rotation state, and its accepted geometry measurements correct the shared homography used to interpret later events from every propeller.
We report lock, the fraction of evaluated time for which an~estimate remains within $2$\,\% of synchronized ground truth, because mean error alone does not show how long a~tracker follows the true rate.
With fixed-rate initialization, leave-one-recording-out evaluation on moving-camera data gives \ours{} $62.9$\,\% lock, compared with $39.8$\,\% for \helixtrack{}~\cite{spetlik_helixtrack_2026} and $22.2$\,\% for \eventpro{}~\cite{chen_count_2026}.
With other settings fixed, allowing four geometry contributors instead of one raises lock from $3.4$\,\% to $62.9$\,\%.
On a~private target--follower dataset with a~different target quadcopter, \ours{} retains the measurement settings selected on moving-camera data and uses a~fixed initial rate. It reaches $95.9$\,\% lock, compared with $56.2$\,\% for \helixtrack{} and $61.1$\,\% for \eventpro{}.
\end{abstract}\vspace{-2em}

\section{Introduction}
\label{sec:intro}

\begin{figure}[t]
\centering
\noindent\resizebox{\linewidth}{!}{\input{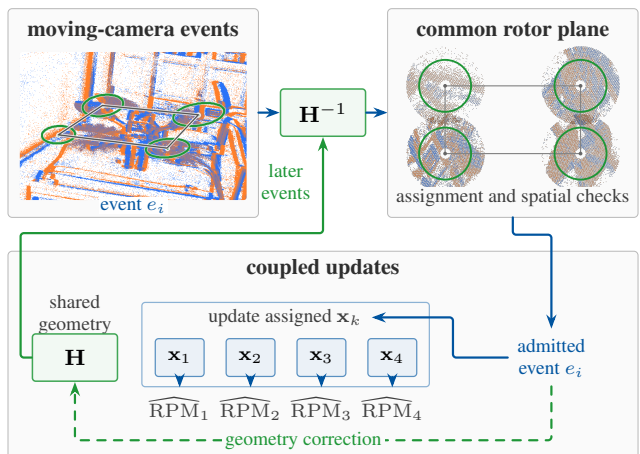}}
\caption{\textbf{Coupled geometry and rotation estimation.} The shared homography $\mathbf H$ maps the rotor plane to the image. The filter maps event $e_i$ by $\mathbf H^{-1}$, assigns it to a~propeller and applies spatial checks. An~admitted event updates its assigned rotation state $\mathbf x_k$, and its accepted geometry measurements correct $\mathbf H$ for later events from every propeller. Each state $\mathbf x_1,\ldots,\mathbf x_4$ contains phase, rate and acceleration and supplies its own shaft-rate estimate $\widehat{\mathrm{RPM}}_1,\ldots,\widehat{\mathrm{RPM}}_4$. Insets integrate $10$~ms from the highest-motion $4$~m recording; color denotes polarity. Green curves show equal rotor-plane rings and their different image projections under the same homography; grey lines join their centres.}\label{fig:overview}%
\end{figure}

We study rotation-rate tracking for four two-blade propellers observed by a~moving event camera. Relative motion between the camera and airframe changes each propeller's image projection while its blades rotate. Recovering the rates requires estimating this changing geometry and the individual rotations from the same events.

Propeller rates set the thrust of a~multicopter, so a~commanded change of its motion starts as a~change of these rates. Cameras already detect and localize drones, also by their propellers~\cite{sanket_evpropnet_2021,stewart_virtual_2022,vrba_marker-less_2020,stewart_drone_2021}, and radar measures propeller rates~\cite{ciattaglia_measuring_2024}. Event cameras~\cite{gallego_event-based_2022} resolve the brightness changes produced by individual blade passages without forming frames.

Event cameras also respond to brightness changes caused by camera motion. Background structure swept through a~propeller's annulus adds events that mix with the blade events, and four nearly identical propellers make event assignment ambiguous. A~rate estimator must therefore separate blade events from background motion before interpreting their timing.\looseness=-1

Interpreting blade events requires knowing where each projected propeller lies. The blades sweep a~circular disc in their own \emph{rotor plane}. Perspective maps this disc to a~moving, generally non-circular image region. We represent blade orientation by a~periodic phase. Mapping an~event to the rotor plane determines its propeller assignment and blade angle. Map errors affect phase and rate estimates.

Existing event-based rate estimators estimate propeller geometry from accumulated events rather than updating it after each event. \helixtrack{}~\cite{spetlik_helixtrack_2026} updates phase per event but refines one propeller's homography from accumulated events. \eventpro{}~\cite{chen_count_2026} estimates each propeller centre from clustered events and its in-plane rate from an~adaptive event batch. Neither method jointly uses the propellers to constrain their common geometry.\looseness=-1

For rigidly mounted, coplanar propellers, one homography maps their fixed rotor-plane layout into the image. Each propeller still has its own phase, rate and acceleration. The resulting estimation problem couples one changing projection to four independent rotations.

We propose \ours{}, a~recursive filter that estimates this homography and the four rotation states through individual-event updates (Fig.~\ref{fig:overview}). Each admitted event corrects its assigned rotation state, and its accepted geometry measurements also correct the shared geometry. That geometry correction changes how later events from every propeller are assigned and interpreted. In contrast to \helixtrack{}~\cite{spetlik_helixtrack_2026}, the coupling acts through a~shared map updated from individual events, while the rotation dynamics remain separate.

Evaluating the resulting rates requires accounting for the two-blade ambiguity. A~two-blade propeller looks unchanged after a~half turn, so an~estimate at twice the reference rate fits the observed events as well as the reference rate does. In addition to mean absolute and relative rate errors, we report the lock percentage: the fraction of evaluated time for which a~reported rate remains within $2$\,\% of synchronized ground truth. Missing output counts as not locked; half- and double-rate estimates fail this test.

In summary, our contributions are:
\begin{enumerate}[label=(\roman*),leftmargin=*,itemsep=1pt,topsep=2pt]
    \item We introduce \ours{}, a~recursive estimator of a~shared projective map and independent propeller rotations. The accepted geometry measurements of each event correct the geometry used to interpret later events across all four propellers. Comparisons quantify the effect of geometry contributors and of the tested Kalman and windowed Gauss--Newton formulations (Secs.~\ref{sec:exp_joint}--\ref{sec:exp_recursion}).
    \item \ours{} reaches the highest lock percentage among the evaluated methods on two datasets with different quadcopters. On the benchmark from~\cite{spetlik_helixtrack_2026}, leave-one-recording-out evaluation with a~fixed initial rate gives $62.9$\,\% lock, compared with $39.8$\,\% for \helixtrack{}~\cite{spetlik_helixtrack_2026} and $22.2$\,\% for \eventpro{}~\cite{chen_count_2026}. With the selected measurement settings unchanged, \ours{} reaches $95.9$\,\% lock on a~private target--follower dataset (Sec.~\ref{sec:exp_comparison}).
\end{enumerate}

\section{Related Work}
\label{sec:related}

Appearance-based event trackers follow Gaussian blobs~\cite{wang_asynchronous_2024}, corners that agree with neighbouring events in space and time~\cite{alzugaray_asynchronous_2018}, or learned templates in event volumes~\cite{messikommer_data-driven_2025,li_3d_2024}. Multi-object methods add learned association and sequence models~\cite{wang_event_2024,wang_mambaevt_2025,sun_reliable_2024,wang_spikemot_2025,zafeiri_event-ecc_2024}. These methods output object positions or feature tracks rather than rotation rates, and two identical blades look the same after half a~turn.

Methods designed specifically for periodic motion instead correlate a~template with later events~\cite{kolar_eeppr_2025,kolar_ee3p_2024}, register consecutive stream slices~\cite{zhao_ev-tach_2024}, estimate a~frequency at each pixel~\cite{pfrommer_frequency_2022,bezick_real-time_2026,spetlik_efficient_2025}, or apply fixed kernels to the pixels around several propellers~\cite{de_araujo_azevedo_event-based_2022,azevedo_multiple_2025}. \eventpro{}~\cite{chen_count_2026} forms event-count and polarity maps, clusters the remaining events to estimate each propeller centre, and selects the in-plane angular rate that best aligns an~adaptive, centre-warped event batch. Its experiments use a~stationary camera, and the method maintains no state for camera translation or perspective changes.

Most event-based aircraft methods detect or localize the airframe~\cite{sanket_evpropnet_2021,vrba_marker-less_2020,mitrokhin_event-based_2018,stewart_drone_2021,stewart_virtual_2022,falanga_dynamic_2020}. \helixtrack{}~\cite{spetlik_helixtrack_2026} instead estimates the rate of one propeller from the spiral pattern formed by its events over time. Its phase filter is updated after every event, while its pose is refined by Gauss--Newton method from accumulated events and radial counts after each half revolution. It processes each propeller independently and does not update shared geometry from individual events.

Innovation gating changes which measurement errors reach a~filter by rejecting those that are large under the current estimate~\cite{or_selection-induced_2026}. Nonlinear updates also make the stored uncertainty too small~\cite{jiang_mitigating_2026}. These concerns motivate our phase-gating and covariance comparisons. We separately compare windowed and per-event geometry updates because they map later events using different geometry estimates.

A~different family of methods estimates motion by maximizing the contrast of accumulated events. These methods warp events by a~candidate motion and optimize the sharpness of their accumulation~\cite{gallego_accurate_2017,gallego_unifying_2018}. They include homographies for moving planes, but require repeated warps of a~window, and an~incorrect motion model also produces sharp accumulations~\cite{shiba_secrets_2024,guo_cmax-slam_2024}. Processing cascades reduce this cost by rejecting events before expensive operations~\cite{mohamed_dynamic_2021,delbruck_frame-free_2008,clerico_retina-inspired_2025}. We also reject events using cheap spatial checks before updating the state with measurements.

Among the propeller-rate methods reviewed, none updates a~shared geometry from individual events while keeping the propeller rates separate.

\section{Method}
\label{sec:method}

Relative motion changes all propeller projections on a~rigid airframe, while their rotation rates differ. \ours{} uses one shared homography and tracks each rotation separately. \helixtrack{}~\cite{spetlik_helixtrack_2026} processes each propeller independently and refines its geometry from event batches. \ours{} applies Kalman geometry updates to individual events. An~accepted observation from one propeller changes how the filter interprets the next event from another (Table~\ref{tab:method_comparison}).

\begin{table}[t]
\centering
\setlength{\tabcolsep}{3pt}
\begin{tabular*}{\linewidth}{@{\extracolsep{\fill}}p{0.24\linewidth}p{0.30\linewidth}p{0.34\linewidth}@{}}
\toprule
 & \helixtrack{}~\cite{spetlik_helixtrack_2026} & \ours{} \\
\midrule
Geometry & \shortstack[l]{one map per\\propeller} & \shortstack[l]{one map for\\all propellers} \\
\shortstack[l]{Geometry\\update} & \shortstack[l]{batched\\Gauss--Newton} & \shortstack[l]{per-event\\Kalman} \\
\shortstack[l]{Rotation\\states} & \shortstack[l]{separate\\per propeller} & \shortstack[l]{separate\\per propeller} \\
\bottomrule
\end{tabular*}
\caption{\textbf{Geometry and rotation updates.} Both methods update phase per event. \ours{} also uses accumulated centre measurements and an~ellipse-based drift observer.}\label{tab:method_comparison}
\end{table}

For $N$ coplanar $B$-blade propellers, the model has six homography parameters and three rotation variables per propeller: phase, rate and acceleration. Our four-propeller evaluation has $6+4\times3=18$ parameters and $B=2$. Initialization requires one ring annotation~\cite{spetlik_efficient_2025}, rotation sign and initial rate per propeller. This section follows an~event through prediction, spatial checks, measurement and update.

\subsection{State and dynamics}
\label{sec:notation}
An~event $e_i=(\mathbf{z}_i,t_i,p_i)$ contains a~pixel location, a~timestamp and a~polarity. The polarity $p_i\!\in\!\{\pm1\}$ records whether that pixel became brighter or darker. The index $k$ identifies a~propeller, $N$ is the propeller count, and $B{=}2$ is the blade count. One homography with six parameters maps the rotor plane into the image:
\begin{equation}\label{eq:homography}
\mathbf{H}(\mathbf{q})=\begin{bmatrix} s\mathbf{R}(\psi) & \mathbf{t}\\ \mathbf{h}_3^\top & 1\end{bmatrix},\quad
\mathbf{q}=[\log s,\,\psi,\,\mathbf{t}^\top,\,\mathbf{h}_3^\top]^\top,
\end{equation}
Optimizing $\log s$ keeps the scale positive and makes equal state increments change it by equal percentages. Applying the inverse initial homography to the N annotated propeller image centres gives fixed rotor-plane centres $\mathbf{c}_k$. We use the propellers' common physical radius as one unit, so $\rho_k\equiv1$. The map for propeller $k$ is then $\mathbf{H}_k=\mathbf{H}\big[\begin{smallmatrix}\rho_k\mathbf{I}&\mathbf{c}_k\\ \mathbf{0}^\top&1\end{smallmatrix}\big]$. Perspective in the homography accounts for apparent size differences across the four image locations; no propeller therefore needs its own scale parameter.

We restrict the linear part of $\mathbf{H}$ to a~scale and rotation instead of estimating independent affine scale and shear. Table~\ref{tab:core_ablations} and Sec.~\ref{sec:suppl_arms} give the controlled comparison with a~full eight-parameter model.

Each propeller has state $\mathbf{x}_k=[\phi_k,\omega_k,\alpha_k]^\top$. Between events, we assume constant acceleration and use a~white-jerk model together with an~additional rate-scaled random walk on angular rate; Sec.~\ref{sec:suppl_config} gives both noise densities. 
Internally, $\phi_k$ is the $B$-fold blade-passage phase and $\omega_k$ is its angular rate in rad\,s$^{-1}$. 

A~$B$-blade propeller looks unchanged after a~physical rotation of $2\pi/B$. In the $B$-fold phase state, equivalent blade appearances are therefore separated by $2\pi$, so we limit the phase variance to that of a~uniform distribution on this circle, $\sigma_{\max}^2=\pi^2/3$. This cap keeps the Gaussian phase approximation within one equivalence interval. Sec.~\ref{sec:suppl_config} explains the variance cap.

We store $N+1$ separate covariance matrices: $\mathbf{P}^k=\operatorname{Cov}(\mathbf{x}_k)$ for each propeller and $\mathbf{P}^q=\operatorname{Cov}(\mathbf{q})$ for the homography. The phase measurement updates only $(\mathbf{x}_k,\mathbf{P}^k)$, with $\mathbf{q}$ held fixed. The active radial, boundary, polarity, signed-crossing, silhouette and centre measurements update only $(\mathbf{q},\mathbf{P}^q)$, with the propeller states held fixed. We therefore store no phase--geometry cross-covariances. Each measurement updates only the relevant small matrix rather than a~full $(6+3N)\times(6+3N)$ matrix. Sec.~\ref{sec:exp_joint} evaluates sharing the homography, while Sec.~\ref{sec:suppl_arms} tests whether neglecting the cross-covariances changes the recovered rates.

\subsection{Event admission}\label{sec:admission}
An~event must be assigned before it updates a~propeller or the homography. We first reject events outside a~padded region around every projected ring, before computing coordinates in the rotor plane. The inverse homography $\mathbf{M}=\mathbf{H}^{-1}$ then maps each remaining event into this plane, where it is assigned to the nearest fixed propeller centre $k^\star$. If the homography maps an~event to or behind the projective horizon, its position in the rotor plane is undefined and we discard it. Assignment in the rotor plane keeps each propeller's identity fixed while its image projection moves and shears.

After assignment, a~radial gate determines whether the event belongs to the propeller. Let $\mathbf{u}$ be the event position after mapping it to the rotor plane, and let $k^\star$ index its assigned propeller, whose fixed rotor-plane centre and normalized radius are $\mathbf{c}_{k^\star}$ and $\rho_{k^\star}=1$. We define local coordinates $\mathbf{l}=(\mathbf{u}-\mathbf{c}_{k^\star})/\rho_{k^\star}$ and radius $r=\|\mathbf{l}\|$, so the annotated ring lies at $r=1$. After every ring is scaled to unit radius, the same nominal thresholds $r_\text{in}$ and $r_\text{out}$ apply to all propellers. The effective outer threshold is clipped separately for each propeller, and the gate accepts $r\in[r_{\text{in}},r_{\text{out},k^\star}]$. We set $r_{\text{out},k^\star}$ to the smallest of $r_{\text{out}}$, the image-boundary limit and half the distance to the nearest centre of another propeller. This prevents the propeller regions from overlapping.

After an~event gap longer than a~threshold duration $\tau_\text{gap}$, we reset phase uncertainty to the variance of a~uniform phase but retain $\omega$ and $\alpha$. During re-acquisition, we temporarily flatten phase-dependent association weights to reduce their sensitivity to phase error.

\subsection{Measurements}\label{sec:rows}
Each admitted event produces one scalar residual for the assigned propeller phase and possibly several scalar residuals for shared geometry. For state $\mathbf y$, each residual $\varepsilon(\mathbf y)$ has Jacobian $\partial\varepsilon/\partial\mathbf y$, base variance $R_{\text{base}}$, and weight $w$, with $R_{\text{eff}}=R_{\text{base}}/w$. Sec.~\ref{sec:suppl_rows} lists each measurement’s residual, Jacobian, support, weight, and parameters.

All phase-dependent measurements use the angular difference between an~event and the predicted blade:
\[
\begin{aligned}
\varepsilon^\phi_i
  &=\wrap{\phi_{k}(t_i)-B\psi-B\,\theta_{\zeta_k}(\mathbf{l}_i)},\\
\theta_{\zeta}(\mathbf{l})
  &=\operatorname{atan2}(\zeta l_y,l_x),
\end{aligned}
\]
where $k=k^\star$ is the propeller assigned to event $i$ and $\zeta\in\{\pm1\}$ is its rotation sign. Here $\phi_k$ is the signed $B$-fold blade phase in the image-oriented gauge of $\mathbf H$, whereas $\theta_{\zeta_k}(\mathbf l_i)$ is measured after back-warping into the rotor plane. Subtracting $B\psi$ therefore expresses $\phi_k$ in the rotor-plane gauge before comparing the two angles. Following \helixtrack{}~\cite{spetlik_helixtrack_2026}, a~ring-distance weight downweights events far from the ring without rejecting them. The optional predicted-blade weight is disabled in the reported configuration.

Unlike geometry, phase does not use innovation gating: no phase measurement is rejected because its error is large under the predicted state. We apply $\varepsilon^\phi_i$ to every event that passes the spatial checks; distance from the ring changes only its weight. Rejecting according to the current estimate would retain errors that already agree with it and change the data presented to the filter.

The event radius $r_i=\|\mathbf l_i\|$ is the normalized distance to the assigned propeller centre, with radial residual $\varepsilon_i^r=r_i-1$. Its Jacobian $\partial\varepsilon_i^r/\partial\mathbf q$ constrains scale and perspective because changing $\mathbf q$ shifts the back-warped position $\mathbf u_i$ and thus $r_i$. We evaluate this over equal-width intervals on both sides of $r_i=1$ to avoid one-sided scale corrections from partially visible rings.

Two soft boundary residuals penalize events inside the modelled inner boundary or outside its outer boundary. Event polarity selects one of two angular offsets in the phase residual, associating positive and negative events with opposing blade edges. These residuals update $\mathbf q$ through their Jacobians; Sec.~\ref{sec:suppl_rows} lists equations and parameters.

After every $2\,000$ admitted events for one propeller, a~centre measurement compares the projected hub with their radially weighted mean position. We omit this measurement when the events form a~one-sided arc, whose mean would be displaced towards its visible side.

Two additional measurements account for the blade's finite width. Signed crossing tests whether event polarity agrees with the direction in which the modelled contour crosses the pixel. Rotating silhouette tests whether the event radius lies within a~phase-dependent blade envelope. The predicted blade outline rotates with phase, so this measurement uses the observed blade edge to correct the geometry.

The phase, radial, boundary, polarity, signed-crossing and silhouette residuals are computed from individual events. The centre measurement is instead formed after $2\,000$ admitted events. A~separate drift observer maintains exponentially decayed event activity and second moments, with time constants given in Sec.~\ref{sec:suppl_config}, and fits one ellipse per propeller. Changes in the fitted centres relative to their slowly varying references provide bounded corrections to the shared translation and to the scale of the four-centre constellation. Sec.~\ref{sec:suppl_arms} isolates the observer's contribution.

\begin{figure*}[t]
\centering
\includegraphics[width=\textwidth]{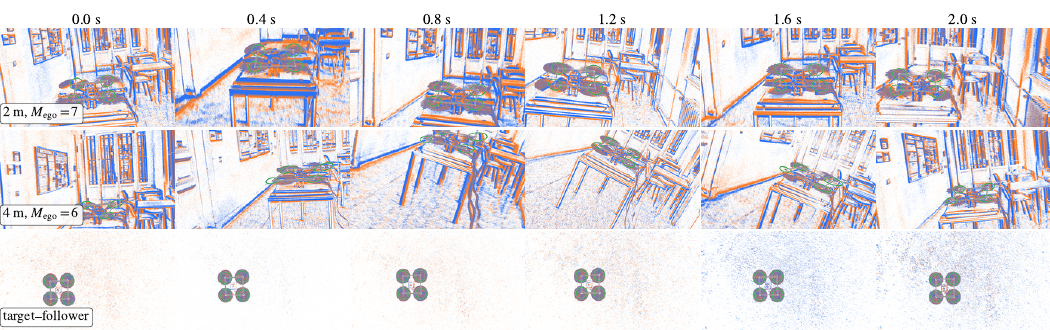}
\caption{\textbf{Event streams and tracked propellers.} Each panel integrates $10$~ms of events; warm and cool colors show the two polarities. Columns begin about $0.5$~s after the fixed evaluation cut of each recording and continue every $0.4$~s. These times were chosen without examining tracker output. The first two rows show the highest camera-motion levels at $2$ and $4$~m. The third row shows the target--follower recording in which the camera travels farthest. Green circles are the propeller rings tracked by \ours{}, and grey lines connect their centres under the shared geometry. Sec.~\ref{sec:suppl_visualization} gives the exact timestamps.}
\label{fig:dataset_progression}\vspace{-0.5em}
\end{figure*}

\subsection{Update and validity}\label{sec:update}
The extended Kalman filter applies each individual-event residual when its event arrives and applies the centre residual after each group of $2\,000$ admitted events. \helixtrack{}~\cite{spetlik_helixtrack_2026} applies a~Kalman update to phase but refines geometry with Gauss--Newton method after each half revolution; \ours{} applies Kalman updates to both phase and geometry.

For geometry residual $\varepsilon_i^q$, let $\mathbf{J}_i^q=\partial\varepsilon_i^q/\partial\mathbf{q}$. The scalar update is
\begin{equation}\label{eq:joseph}
\begin{aligned}
&S_i\!=\!\mathbf{J}_i^q\mathbf{P}^q(\mathbf{J}_i^q)^\top\!+\!R_{\text{eff}},\hspace{0.8em}
\mathbf{K}_i\!=\!\frac{\mathbf{P}^q(\mathbf{J}_i^q)^\top}{S_i},\hspace{0.8em}
\mathbf{q}^{\mathrm{cand}}\!=\!\mathbf{q}-\mathbf{K}_i\varepsilon_i^q,\\
&\mathbf{P}^{q,\mathrm{cand}}
=(\mathbf{I}-\mathbf{K}_i\mathbf{J}_i^q)\mathbf{P}^q
(\mathbf{I}-\mathbf{K}_i\mathbf{J}_i^q)^\top
+\mathbf{K}_iR_{\text{eff}}\mathbf{K}_i^\top.
\end{aligned}
\end{equation}
The Joseph covariance form limits numerical loss of symmetry and positive semi-definiteness.

We accept a~geometry residual only if its normalized innovation satisfies
\[
\nu_i=\frac{(\varepsilon_i^q)^2}{S_i}\le\tau_{\mathrm{NIS}},
\qquad \tau_{\mathrm{NIS}}=16.
\]
One geometry update changes the back-warp used to assign later events from all propellers, so this gate limits the effect of an~inconsistent residual on the shared state. We do not gate phase residuals. Phase is periodic, and after an~event gap or an~incorrect prediction the necessary correction is large; rejecting such a~correction according to the current phase estimate keeps the wrong phase.

Geometry candidates must be finite and satisfy the scale, translation, perspective, determinant and condition-number bounds in Sec.~\ref{sec:suppl_validity}. The projective denominator must be positive at all four corners of the rotor-plane box containing the propeller rings. Its linearity guarantees positivity throughout the box. Rejected candidates leave the state and covariance unchanged. These checks exclude the stated numerical and projective failures but do not establish pose accuracy.

Let $t_k$ be the time at which propeller state $k$ was last propagated. Between events, the state predicts $\hat\omega_k(t)=\omega_k+\alpha_k\max(t-t_k,0)$. This prediction provides a~rate at any requested time without waiting for another event. For the evaluated fixed configuration $N=4$, each accepted event requires $\mathcal{O}(1)$ work.%

\section{Experiments}\label{sec:experiments}

We evaluate rate tracking against synchronized tachometer references on two datasets. As in \helixtrack{}~\cite{spetlik_helixtrack_2026}, we report lock, with mean errors and coverage. The homography controls event assignment and phase, so its errors affect the recovered rates. We do not score homography error.

\subsection{Datasets}\label{sec:exp_dataset}

Figure~\ref{fig:dataset_progression} shows representative event streams from the two datasets, which use different quadcopters.

\noindent\textbf{\tquadcopterego{} (\tquadcopteregoshort{}).} This moving-camera benchmark~\cite{spetlik_helixtrack_2026} contains thirteen recordings of a~table-mounted quadcopter with a~$630$~mm diagonal between opposite motor centres, viewed by a~moving Metavision EVK4HD at $1280\times720$, giving 52 propeller tracks. Seven recordings are captured from $2$~m and six from $4$~m. At each distance, the sequence index gives the camera-motion level $M_\text{ego}$. An~infrared tachometer records each propeller's blade-pass rate on the camera clock. We divide these rates by the known blade count and report shaft RPM. We use the benchmark recordings, temporal cuts and synchronized reference exactly as provided.

Nine gaps in the $2$~m, $M_\text{ego}=4$ event stream total $664$~ms and meet the input-only damage criterion in Sec.~\ref{sec:suppl_streams}. We therefore keep it outside configuration selection and report it as a~stress test. Including this recording does not change the method ordering.

\noindent\textbf{Target--follower.} This private dataset contains three recordings of a~quadcopter with a~$500$~mm diagonal between opposite motor centres, flying above textured ground. An~event camera on a~following quadcopter views it from above. The four propeller tracks per recording give $803.6$ propeller-seconds for evaluation. An~optical gate directly supplies the synchronized shaft-rate reference. We divide the trackers' blade-pass estimates by the known blade count $B$ before comparison. Sec.~\ref{sec:suppl_scoring} reports its timing and pulse-consistency checks. No second sensor independently verifies this reference.

\begin{table*}[t]
\centering
\renewcommand{\arraystretch}{0.88}
\setlength{\tabcolsep}{1.2pt}
\begin{tabular*}{\textwidth}{@{\extracolsep{\fill}}c c *{12}{>{\centering\arraybackslash}p{0.059\textwidth}}@{}}
\toprule
& & \multicolumn{4}{c}{\ours{}} & \multicolumn{4}{c}{\helixtrack{}~\cite{spetlik_helixtrack_2026}} & \multicolumn{4}{c}{\eventpro{}~\cite{chen_count_2026}} \\
\cmidrule(lr){3-6} \cmidrule(lr){7-10} \cmidrule(lr){11-14}
$d$ & $M_\text{ego}$ & $T_\text{lock}$ & \% & MAE & cov\% & $T_\text{lock}$ & \% & MAE & cov\% & $T_\text{lock}$ & \% & MAE & cov\% \\
\midrule
\multirow{6}{*}{$2$\,m} & 1 & \textbf{94.6} & \textbf{82} & \textbf{43} & 100 & 93.6 & 81 & 47 & 100 & 51.2 & 45 & 1000 & 100 \\
 & 2 & \textbf{89.3} & \textbf{81} & \textbf{44} & 100 & 88.7 & 80 & 45 & 100 & 38.2 & 34 & 1348 & 100 \\
 & 3 & \textbf{90.3} & \textbf{81} & \textbf{56} & 100 & 75.2 & 68 & 33661 & 100 & 34.2 & 31 & 1256 & 99 \\
 & 5 & \textbf{53.4} & \textbf{46} & \textbf{1565} & 100 & 5.9 & 5 & 3825 & 92 & 6.5 & 6 & 2119 & 45 \\
 & 6 & \textbf{43.8} & \textbf{39} & 7653 & 100 & 2.7 & 2 & 4347 & 100 & 2.1 & 2 & \textbf{2330} & 31 \\
 & 7 & \textbf{17.9} & \textbf{16} & 2819 & 100 & 2.2 & 2 & 427951 & 100 & 0.9 & 1 & \textbf{2390} & 20 \\
\midrule
\multirow{6}{*}{$4$\,m} & 1 & 69.2 & 60 & 1174 & 100 & \textbf{71.6} & \textbf{62} & 1045 & 100 & 55.2 & 48 & \textbf{667} & 100 \\
 & 2 & 68.2 & 60 & 1170 & 100 & \textbf{68.6} & \textbf{61} & 1213 & 100 & 58.7 & 52 & \textbf{583} & 100 \\
 & 3 & \textbf{61.8} & \textbf{55} & 1392 & 100 & 50.3 & 45 & 2213 & 100 & 44.6 & 40 & \textbf{1016} & 100 \\
 & 4 & \textbf{94.9} & \textbf{83} & \textbf{81} & 100 & 74.2 & 65 & 1138 & 100 & 2.7 & 2 & 1950 & 19 \\
 & 5 & \textbf{87.3} & \textbf{80} & \textbf{152} & 100 & 1.5 & 1 & 4784 & 100 & 6.1 & 6 & 1922 & 40 \\
 & 6 & \textbf{83.2} & \textbf{72} & \textbf{451} & 100 & 4.8 & 4 & 3735 & 100 & 1.2 & 1 & 2643 & 23 \\
\midrule
\multicolumn{2}{l}{\textbf{all 48 tracks}} & \textbf{854.0} & \textbf{63} & \textbf{1383} & 100 & 539.4 & 40 & 40334 & 99 & 301.5 & 22 & 1602 & 64 \\
\bottomrule
\end{tabular*}
\caption{\textbf{Timestamped Quadcopter with Egomotion benchmark}: 12 recordings $\times$ 4 propellers, $1356.9$ propeller-seconds. Every tuned setting was selected without the recording it scores. \ours{} and \helixtrack{}~\cite{spetlik_helixtrack_2026} start every propeller at $4\,750$ shaft RPM. Every method is scored at every timestamp. \% is the lock percentage (Eq.~\ref{eq:locktime}), $T_\text{lock}$ is its duration in propeller-seconds, MAE is the mean absolute shaft-RPM error and cov\,\% is coverage, the percentage of evaluation timestamps with an~estimate. Higher is better except for MAE. Bold marks the best lock and error values in each row; coverage is descriptive. The damaged $2$\,m, $M_\text{ego}\!=\!4$ recording is reported separately in Sec.~\ref{sec:suppl_streams}.}\label{tab:benchmark_ego}\vspace{-1em}
\end{table*}

\begin{table}[t]
\centering
\setlength{\tabcolsep}{5pt}
\begin{tabular*}{\linewidth}{@{\extracolsep{\fill}}l r r r r@{}}
\toprule
Method & $T_\text{lock}$ & \% & MAE & cov\% \\
\midrule
\ours{} & \textbf{770.4} & \textbf{95.9} & \textbf{68.4} & 100.0 \\
\eventpro{}~\cite{chen_count_2026} & 490.9 & 61.1 & 115.0 & 100.0 \\
\helixtrack{}~\cite{spetlik_helixtrack_2026} & 451.7 & 56.2 & 1767.1 & 98.0 \\
\bottomrule
\end{tabular*}
\caption{\textbf{Private target--follower dataset}: 3 recordings $\times$ 4 propellers, $803.6$ propeller-seconds. \ours{} and \helixtrack{}~\cite{spetlik_helixtrack_2026} are initialized at $9\,500$ shaft RPM. The \ours{} and \eventpro{}~\cite{chen_count_2026} configurations were selected on the TQE recordings and applied unchanged. Lock is the percentage of evaluated propeller-time; $T_\text{lock}$ is the corresponding duration. MAE is mean absolute shaft-RPM error; cov\,\% is coverage, the percentage of evaluation timestamps with an~estimate.}\label{tab:benchmark_tgt}\vspace{-1em}
\end{table}

\subsection{Measures and protocol}

\label{sec:exp_metric}
We use the lock percentage together with mean error and coverage. Every $5$~ms, we test whether a~reported propeller rate is within $2$\,\% of synchronized ground truth. Let $\ell_{s,k}(t)$ denote this test for propeller $k$ on recording grid $\mathcal{T}_s$. The lock percentage is
\begin{equation}
\begin{aligned}
\ell_{s,k}(t)&=
\begin{cases}
\mathbf{1}\!\left[\left|\dfrac{\widehat{\mathrm{RPM}}_{s,k}(t)}{\mathrm{RPM}^\text{gt}_{s,k}(t)}-1\right|\le\tau_\text{lock}\right],
& \text{\parbox{5.5em}{if an~estimate exists,}}\\
0, & \text{otherwise},
\end{cases}\\
P_\text{lock}&=100\,\frac{\sum_{(s,k)}\sum_{t\in\mathcal{T}_s}\ell_{s,k}(t)}{\sum_{(s,k)}|\mathcal{T}_s|}.
\end{aligned}
\label{eq:locktime}
\end{equation}
where $\tau_\text{lock}=0.02$ for every method. We retain the $2$\,\% lock criterion used by \helixtrack{}~\cite{spetlik_helixtrack_2026}; the supplement reports the same trajectories at $1$ and $5$\,\%. The relative tolerance applies equally throughout the evaluated rate range and rejects half-rate and double-rate estimates. A~missing estimate is counted as unlocked. For reproducibility, the tables also give the corresponding duration $T_\text{lock}$ in propeller-seconds. We report mean absolute shaft-RPM error (MAE), mean absolute relative error (MArE) and coverage. Coverage is the percentage of evaluation timestamps for which the method reports an~estimate. Because MAE and MArE average only the emitted samples, they must be read with coverage.

\label{sec:exp_protocol}
All methods use the same time intervals/timestamps, initial ring annotations, and total scoring duration. \eventpro{}~\cite{chen_count_2026} does not keep annotated centres; it re-estimates them from events.
Sec.~\ref{sec:suppl_scoring} describes how we checked the scorer.

Our \eventpro{}~\cite{chen_count_2026} reimplementation follows the published method. We run public \helixtrack{} code~\cite{spetlik_helixtrack_2026} per propeller and select ring bounds from its published grid.

We use nested leave-one-recording-out evaluation on the 12 TQE recordings. Each fold uses the other 11 recordings to tune all three methods and scores the omitted recording once. \ours{} and \helixtrack{}~\cite{spetlik_helixtrack_2026} start at $4\,750$ shaft RPM throughout selection and evaluation. We aggregate the 12 test scores. For target--follower, we select \ours{} and \eventpro{}~\cite{chen_count_2026} settings on all 12 TQE recordings and retain \helixtrack{}'s public configuration. Measurement settings remain unchanged, while \ours{} and \helixtrack{} start at $9\,500$ shaft RPM (Secs.~\ref{sec:suppl_selection} and~\ref{sec:suppl_arms}).

We omit complete recordings because \ours{} connects the four propellers through one homography. The propeller-wise four-fold protocol reported with \helixtrack{}~\cite{spetlik_helixtrack_2026} retains all camera motions in every fold. Sec.~\ref{sec:suppl_propeller_cv} reproduces it.

\ours{} and \helixtrack{}~\cite{spetlik_helixtrack_2026} internally store blade-pass rate. We divide every blade-pass output and the TQE reference by $B=2$. Target--follower already provides shaft rate. Rates and absolute errors are in shaft RPM (Sec.~\ref{sec:suppl_arms}).

\subsection{Comparison with prior methods}\label{sec:exp_comparison}
On 12 TQE recordings, \ours{} achieves $62.9$\,\% lock, $1383$~RPM shaft-rate MAE, and $99.9$\,\% coverage. The public \helixtrack{} implementation~\cite{spetlik_helixtrack_2026} achieves $39.8$\,\% lock and $99.2$\,\% coverage. The comparison includes shared geometry and per-event geometry updates; Secs.~\ref{sec:exp_joint}--\ref{sec:exp_recursion} compare contributor sets and update formulations.

\eventpro{}~\cite{chen_count_2026} reaches $22.2$\,\% lock, with a~shaft-rate MAE of $1602$~RPM and $64.5$\,\% coverage. When its internal checks do not find a~reliable rate, it reports no value. These gaps count as unlocked in the common score.

\begin{figure*}[t]\centering
\includegraphics[width=\linewidth]{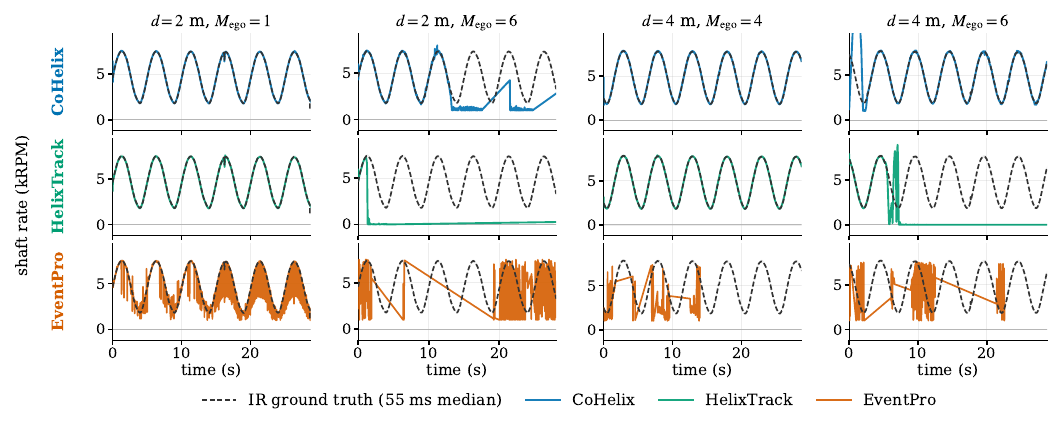}\vspace{-1em}
\caption{\textbf{Shaft rate over time} for one propeller in four recordings spanning the camera-motion range. Rows show \ours{}, \helixtrack{}~\cite{spetlik_helixtrack_2026} and \eventpro{}~\cite{chen_count_2026} together with the same dashed ground-truth reference, filtered by a~centred $55$~ms median for display only; scoring uses the synchronized trace without filtering. The vertical axis ends at $1.25\times$ the largest filtered reference rate, so larger estimates are clipped. Gaps in the \eventpro{} row are samples for which that method does not report.}\label{fig:traces}\vspace{-1em}
\end{figure*}

Figure~\ref{fig:traces} shows representative traces for all three methods across the camera-motion range, including these gaps.

Testing the reported advantage across $100\,000$ alternative mixtures of the available recordings gives 95\,\% intervals of $[9.6,38.8]$ percentage points over \helixtrack{}~\cite{spetlik_helixtrack_2026} and $[27.4,54.4]$ over \eventpro{}~\cite{chen_count_2026} (Sec.~\ref{sec:suppl_uncertainty}).

On the target--follower recordings, \ours{} reaches $95.9$\,\% lock, $34.8$ percentage points above \eventpro{}~\cite{chen_count_2026} ($61.1$\,\%) and $39.7$ points above the public \helixtrack{} implementation~\cite{spetlik_helixtrack_2026} ($56.2$\,\%). It has higher lock on each of the three recordings. Sec.~\ref{sec:suppl_uncertainty} gives uncertainty from whole-recording resampling.

\ours{} has higher lock than the public \helixtrack{} implementation~\cite{spetlik_helixtrack_2026} on 10 of the 12 TQE recordings; \helixtrack{} is higher at the two lowest camera-motion levels recorded from $4$~m. \ours{} has higher lock than \eventpro{}~\cite{chen_count_2026} on all 12 recordings. Tightening the lock tolerance to $1$\,\% or widening it to $5$\,\% preserves the aggregate ordering of all three methods on both datasets (Sec.~\ref{sec:suppl_thresholds}).

For two short \helixtrack{}~\cite{spetlik_helixtrack_2026} runs, counting missing suffixes as locked preserves the ordering (Sec.~\ref{sec:suppl_helixtrack_prefix}).{\widowpenalty=10000\par}

\subsection{Shared geometry ablation}\label{sec:exp_joint}
For each tested recording, we use the configuration selected on the other 11 recordings and change only which propellers contribute to the shared geometry. All four rate states remain active. We average every nonempty subset of equal size so that each propeller contributes equally.

Allowing one, two, three and four propellers to contribute to the shared geometry gives $3.4$\,\%, $17.7$\,\%, $50.2$\,\% and $62.9$\,\% lock (Table~\ref{tab:core_ablations}). More contributors supply observations at more image locations and more geometry measurements. This comparison measures their combined benefit.

Four rings at different image locations constrain the same six homography parameters. With one ring, a~change in homography rotation, together with a~translation that keeps its centre fixed and a~phase shift that preserves the blade angle, leaves the observations of that ring unchanged. Observations from other ring centres constrain this shared motion, while their phase states remain independent. This is not a~test of the best tracker designed for one propeller; Sec.~\ref{sec:exp_comparison} compares with \helixtrack{}~\cite{spetlik_helixtrack_2026}. Sec.~\ref{sec:suppl_arms} lists every setting that both versions share.

We also compare one and four contributors under four equal geometry-event caps, with the ellipse observer disabled. Both the average single-contributor and four-contributor conditions reach at most $2.9$\,\% lock, versus $62.3$\,\% for the unbudgeted control. Realized event and scalar-measurement counts differ, so this diagnostic does not isolate the benefit of spatial support (Sec.~\ref{sec:suppl_budgets}).

\subsection{Geometry updates and stored uncertainty}\label{sec:exp_recursion}
Each comparison in Table~\ref{tab:core_ablations} has its own control. Sec.~\ref{sec:suppl_arms} gives the absolute results.

We compare per-event Kalman geometry updates with damped Gauss--Newton updates over windows of $50$, $100$ or $200$~$\mu$s. For each tested recording, the other 11 recordings select the window. Both versions update phase after every event. The per-event filter improves lock by $23.1$ percentage points over the selected windows, with a~95\,\% interval of $[3.7,43.5]$ points. See Sec.~\ref{sec:suppl_arms} for each recording.

The homography controls where the next event is mapped and whether it passes the spatial checks. Per-event updates correct that map before the next event arrives, whereas the windowed alternative delays its geometry-measurement corrections until the window closes. The comparison changes the solver and its uncertainty treatment as well as the update schedule. It supports the per-event filter over the tested windowed formulation (Table~\ref{tab:core_ablations}).

\begin{table}[t]
\centering
\setlength{\tabcolsep}{4pt}
\renewcommand{\arraystretch}{1.08}
\begin{tabular}{@{}>{\raggedright\arraybackslash}p{0.54\linewidth}@{\hspace{0.04\linewidth}}w{r}{0.16\linewidth}@{\hspace{0.04\linewidth}}w{r}{0.22\linewidth}@{}}
\toprule
Comparison & Change [points] & 95\,\% interval \\
\midrule
four contributors vs. one & $+59.6$ & $[47.8,70.1]$ \\
per-event Kalman vs. selected Gauss--Newton window & $+23.1$ & $[3.7,43.5]$ \\
correlations retained vs. cleared & $-2.1$ & $[-13.8,6.8]$ \\
restricted vs. full homography & $+3.9$ & $[-6.5,13.4]$ \\
\bottomrule
\end{tabular}
\caption{\textbf{Effects of four design choices} on the 12 TQE recordings. Each row compares the two named versions under the same evaluation conditions. Change is the lock percentage of the first version minus that of the second; positive values favour the first. Intervals resample complete recordings. Sec.~\ref{sec:suppl_arms} gives the absolute results.}\label{tab:core_ablations}%
\end{table}

Retaining correlations between uncertainty in phase and geometry changes lock by $-2.1$ percentage points. The 95\,\% interval of $[-13.8,6.8]$ points does not establish a~benefit, so we retain the smaller block representation.

Phase and geometry updates from the same event use the same event coordinates and predicted state. A~deterministic rule that does not examine residuals selects one active measurement family per event, with its weight scaled to preserve the expected total weight. Applying all measurements improves lock by $1.9$ percentage points over this control, with a~95\,\% interval of $[-3.3,6.4]$ points. This test does not establish a~difference. We retain the updates for tracking but do not interpret internal uncertainty as calibrated confidence (Sec.~\ref{sec:suppl_event_dependence}).{\widowpenalty=10000\par}

\subsection{Calibration sensitivity}\label{sec:exp_init}\label{sec:exp_ablation}
We vary the supplied ring centres and radii while retaining the selected measurement settings. Centre offsets of $5$\,\% of the mean annotated radius give $61.0$\,\% TQE and $88.4$\,\% target--follower lock. These scores average four direction realizations per recording. Increasing all supplied radii by $5$\,\% reduces target--follower lock from $95.9$\,\% to $65.7$\,\%, with the loss concentrated in one recording. Both changes also alter the fixed layout, so this diagnostic measures sensitivity to supplied calibration under the selected settings (Sec.~\ref{sec:suppl_annotations}).

\subsection{Runtime and diagnostic tests}\label{sec:exp_runtime}
We time \ours{} and the public \helixtrack{} implementation~\cite{spetlik_helixtrack_2026} end to end on one CPU thread. \ours{} returns rates for all four propellers in one run; for \helixtrack{}, we sum four sequential per-propeller runs. The geometric-mean four-rate real-time factor (recording duration divided by processing time) is $0.0306$ versus $0.0189$ on TQE and $0.0478$ versus $0.0593$ on target--follower. \ours{} is faster on TQE and \helixtrack{} on target--follower, but neither runs in real time (Sec.~\ref{sec:suppl_runtime}).

\subsection{Scope of the evaluation}\label{sec:exp_scope}
All methods start from annotated rings, and \ours{} and \helixtrack{}~\cite{spetlik_helixtrack_2026} also receive rotation signs, so we evaluate rate tracking after localization rather than ring detection. The shared homography assumes rigid, coplanar propellers. Flex, non-coplanarity and relative motion between propeller hubs require richer models; motion of the whole rigid airframe preserves the shared-plane assumption.

Both datasets use rate references independent of the event camera. Leave-one-recording-out evaluation tests transfer among TQE recordings. Target--follower uses the same IMX636 sensor, a~moving camera and a~smaller moving target ($500$ instead of $630$~mm motor diagonal).

\section{Conclusion}\label{sec:conclusion}

\ours{} separates shared image motion from independent propeller rotations through one homography and four phase states. With fixed-rate initialization, it reaches $62.9$\,\% lock on moving-camera data and $95.9$\,\% on target--follower. Both exceed the corresponding \helixtrack{}~\cite{spetlik_helixtrack_2026} and \eventpro{}~\cite{chen_count_2026} lock percentages.

Additional geometry contributors improve tracking through combined spatial support and measurement quantity. Per-event Kalman geometry updates also exceed the tested windowed Gauss--Newton formulation by $23.1$ percentage points. This comparison includes differences in solver, uncertainty treatment and update schedule.

\noindent\textbf{Limitations.} The method assumes localized rings and rigid, coplanar propellers. With fixed measurement settings, a~$5$\,\% increase of the supplied ring radii reduces target--follower lock from $95.9$\,\% to $65.7$\,\%. The evaluation covers two tracked quadcopters and one sensor type, so it does not test other cameras, airframes or scenes.

\noindent\textbf{Acknowledgements.} The research reported in this paper has been partly funded by BMK, BMAW, and the State of Upper Austria in the frame of the SCCH competence center INTEGRATE [Project 1.6 TFI (Transferable Intelligence)] part of the FFG COMET Competence Centers for Excellent Technologies Programme, and by the Czech Science Foundation grant 25-15993S. %

\noindent\textbf{Use of generative AI.} OpenAI Codex assisted with manuscript drafting, experiment automation, analysis and presentation. The authors developed the research questions and method. They take full responsibility for the manuscript and for review and verification of AI-assisted work.

{
    \small
    \bibliographystyle{ieeenat_fullname}
    \bibliography{references}
}

\clearpage
\setcounter{page}{1}
\renewcommand{\thepage}{S\arabic{page}}
\maketitlesupplementary

\newcommand{\beginsupplement}{ %
  \setcounter{figure}{0}       %
  \setcounter{table}{0}        %
  \setcounter{equation}{0}     %
  \setcounter{section}{0}      %
  \renewcommand{\thefigure}{{S}\arabic{figure}}%
  \renewcommand{\thetable}{{S}\arabic{table}}%
  \renewcommand{\theequation}{{S}\arabic{equation}}%
  \renewcommand{\thesection}{{S}\arabic{section}}%
  \renewcommand{\theHfigure}{S.\arabic{figure}}%
  \renewcommand{\theHtable}{S.\arabic{table}}%
  \renewcommand{\theHequation}{S.\arabic{equation}}%
  \renewcommand{\theHsection}{S.\arabic{section}}%
}
\beginsupplement

\section{Measurement model}\label{sec:suppl_rows}

We describe the event measurements used by the filter and the alternatives that we tested but did not retain.

Each measurement has a~scalar residual in the ``predicted minus measured'' direction, so its mean update is $\mathbf{x}\leftarrow\mathbf{x}-\mathbf{K}\,\varepsilon$. A~phase measurement for propeller $k$ uses $\mathbf{C}=[\,1\ \ 0\ \ 0\,]$ on $\mathbf{x}_k=[\phi_k,\omega_k,\alpha_k]^\top$. The remaining measurements act on the geometry state $\mathbf{q}=[\log s,\ \psi,\ t_x,\ t_y,\ h_{31},\ h_{32}]^\top$ with covariance $\mathbf{P}^q$. For a~Jacobian $\mathbf{J}\in\mathbb{R}^{1\times6}$, weight $w$ and base variance $R_\text{base}=1$,
\[
R_\text{eff}=\frac{R_\text{base}}{w},\quad
\boldsymbol\pi=\mathbf{P}^{q}\mathbf{J}^\top,\quad
S=\mathbf{J}\boldsymbol\pi+R_\text{eff},\quad
\nu=\frac{\varepsilon^{2}}{S},
\]
we discard the measurement if $w<10^{-6}$ or if its normalized innovation squared exceeds the gate, $\nu>16$. Otherwise, $\mathbf{K}=\boldsymbol\pi/S$ and the Joseph update
$\mathbf{P}^{q}\leftarrow\mathbf{P}^{q}-\mathbf{K}\boldsymbol\pi^\top-\boldsymbol\pi\mathbf{K}^\top+S\,\mathbf{K}\mathbf{K}^\top$
is applied only if the candidate mean $\mathbf{q}-\mathbf{K}\varepsilon$ passes the geometry checks in Sec.~\ref{sec:suppl_validity}.

Several measurements use the same ring kernel, von Mises weight and Huber weight,
\[
w_\text{ring}(r)=\mathrm{e}^{-\frac{1}{2}\left(\frac{r-1}{\sigma_\text{ring}}\right)^{2}},\quad
w_\text{vm}(\varepsilon^\phi)=\mathrm{e}^{\kappa_\text{eff}(\cos\varepsilon^\phi-1)},
\]
\[
w_\text{Huber}(\varepsilon;c)=\min\!\Big\{1,\ \frac{c}{|\varepsilon|}\Big\},
\]
with $\sigma_\text{ring}=0.25$, $\kappa=22$, $c_\phi=0.6$ and $c_r=0.3$. During post-gap re-acquisition, $\kappa_\text{eff}=0.15\,\kappa$; otherwise $\kappa_\text{eff}=\kappa$. We write $\operatorname{softplus}_\tau(d)=\tau\log(1+\mathrm{e}^{d/\tau})$ and $\sigma(z)=(1+\mathrm{e}^{-z})^{-1}$ for the softplus barrier and its derivative.

For admitted event $i$ on propeller $k$, let $\mathbf{l}_i=(l_x,l_y)$ be its local coordinates and $r_i=\|\mathbf{l}_i\|$. Its wrapped helical-phase residual is
\begin{equation}\label{eq:suppl_phase_residual}
\varepsilon^\phi_i=\wrap{\phi_k-B\psi-B\,\theta_i},\qquad
\theta_i=\operatorname{atan2}(\zeta_k l_y,\ l_x),
\end{equation}
where $\wrap{a}=\mathrm{mod}(a+\pi,2\pi)-\pi$ and $\zeta_k\in\{\pm1\}$ is the rotation sign.

Geometry measurements use the $1\times6$ phase Jacobian $\mathbf{J}^{\phi}$ of $-B(\psi+\theta_i)$ and radial Jacobian $\mathbf{J}^{r}$ of $r_i$. With $\mathbf{M}=\mathbf{H}^{-1}$ and back-warped point $\mathbf{u}_i$,
\[
\mathbf{m}=-\begin{bmatrix}\mathbf{u}_i\\1\end{bmatrix},\qquad
\mathbf{w}_a=\begin{bmatrix}M_{1a}\\M_{2a}\end{bmatrix}-M_{3a}\,\mathbf{u}_i,
\qquad
\frac{\partial\mathbf{u}_i}{\partial h_{ab}}=m_b\,\mathbf{w}_a .
\]
Let $\mathbf{a}_i=\frac{B}{\rho_k r_i^{2}}(-\zeta_k l_y,\zeta_k l_x)$ and
$\mathbf{b}_i=\frac{1}{\rho_k r_i}(l_x,l_y)$. With
$\varsigma_a=\mathbf{a}_i\!\cdot\!\mathbf{w}_a$ and
$\varkappa_a=\mathbf{b}_i\!\cdot\!\mathbf{w}_a$, the gradients are
\[
G^{\phi}_{ab}=-m_b\,\varsigma_a,\qquad G^{r}_{ab}=m_b\,\varkappa_a ,
\]
where $m_3=-1$ makes the two translation entries exactly $G^{\phi}_{a3}=+\varsigma_a$ and $G^{r}_{a3}=-\varkappa_a$. Projecting an~entry gradient $\mathbf{G}$ onto $\mathbf{q}$ gives
\[
\begin{aligned}
\partial_{\log s}&=G_{11}h_{11}+G_{12}h_{12}+G_{21}h_{21}+G_{22}h_{22},\\
\partial_{\psi}&=G_{11}h_{12}-G_{12}h_{11}+G_{21}h_{11}+G_{22}h_{12},
\end{aligned}
\]
$\partial_{t_x}=G_{13}$, $\partial_{t_y}=G_{23}$, $\partial_{h_{31}}=G_{31}$, $\partial_{h_{32}}=G_{32}$. The radial Jacobian $\mathbf{J}^{r}$ is the resulting projection of $\mathbf{G}^{r}$. For $\mathbf{J}^{\phi}$, we project $\mathbf{G}^{\phi}$ and subtract $B$ from its $\psi$ entry to account for the explicit $-B\psi$ term in Eq.~(\ref{eq:suppl_phase_residual}). Finite-difference tests verify both Jacobians.

\paragraph{Phase update (R1).} Every admitted event updates the phase without an~innovation gate. The residual is $\varepsilon=\varepsilon^\phi_i$, the Jacobian is $\mathbf{C}=[\,1\ 0\ 0\,]$, and the weight enters as an~information scale,
\[
\begin{aligned}
\mathcal{I}&=\max\!\big\{10^{-3},\ \kappa w_\text{ring}(r_i)w_\text{coh}\big\},\\
R_\text{eff}&=R_\phi/\mathcal{I},\qquad R_\phi=1 .
\end{aligned}
\]
The coherence factor is $w_\text{coh}=\max\{0.02,\,w_\text{vm}(\varepsilon^\phi_i)\}$ when the von Mises phase gate is enabled and $w_\text{coh}=1$ otherwise. The reported configuration does not use this gate, so the phase update has neither an~NIS gate nor an~angular gate. After each update, we limit $|\alpha_k|\le\alpha_\text{max}$ and $|\omega_k|\in[\omega_\text{min},\omega_\text{max}]$ while preserving its sign. If either limit is reached, we set $\alpha_k\leftarrow0$.

We next define the event-based geometry measurements.

\paragraph{Phase-based geometry update (R2).} We do not use this measurement in the reported configuration. Its residual is $\varepsilon=\varepsilon^\phi_i$, its Jacobian is $\mathbf{J}^{\phi}$ and its weight is
$w=\lambda_\phi\,w_\text{ring}(r_i)\,w_\text{vm}(\varepsilon^\phi_i)\,w_\text{Huber}(\varepsilon^\phi_i;c_\phi)$.
Disabling this row removes its phase residual from the geometry update. Other geometry rows remain phase-dependent.

\paragraph{Radial alignment (R3).}
The residual is $\varepsilon=r_i-1$, the Jacobian is $\mathbf{J}^{r}$ and the weight is
$w=\lambda_r\,w_\text{ring}(r_i)\,w_\text{vm}(\varepsilon^\phi_i)\,w_\text{Huber}(r_i-1;c_r)$.
Here $\lambda_r=2.000$. We use the widest symmetric interval around the ring, $r_i\in[1-\Delta,1+\Delta]$, where $\Delta=\min\{r_{\text{out},k}-1,\ 1-r_\text{in}\}$. If $\Delta\le0$, we omit the measurement. Without this symmetry, events from only one side bias the shared scale.

\paragraph{Annulus boundaries (R4).} We constrain each boundary with $\lambda_\text{band}=4.000$. With $d_\text{in}=b_\text{in}-r_i$ and $d_\text{out}=r_i-b_{\text{out},k}$,
\[
\varepsilon=\operatorname{softplus}_\tau(d),\quad
\mathbf{J}=\mp\,\sigma(d/\tau)\,\mathbf{J}^{r},\quad
w=\lambda_\text{band}w_\text{vm},
\]
the sign being $-$ for the inner and $+$ for the outer barrier. Here $b_\text{in}=0.3$, $b_\text{out}=1.3$ and $\tau=0.05$; with gate following, $b_{\text{out},k}=r_{\text{out},k}-2\tau$ and \emph{both} barriers are disabled when $b_{\text{out},k}\le1$, \ie{} when no outward gradient exists.

\paragraph{Event polarity (R5).} Signed polarity $p_i=\pm1$ provides a~second phase observation with $\lambda_\text{pol}=0.500$. With offsets $o_+=+0.25$, $o_-=-0.25$ and $\tilde\varepsilon=\wrap{\varepsilon^\phi_i-o_{p_i}}$,
\[
\varepsilon=2\sin(\tilde\varepsilon/2),\qquad \mathbf{J}=\cos(\tilde\varepsilon/2)\,\mathbf{J}^{\phi},
\]
\[
w=\tfrac{1}{2}\lambda_\text{pol}\,\kappa_p\,G(r_i)^{2}\,\mathrm{e}^{\kappa_p(\cos\tilde\varepsilon-1)}\min\{1,1/|\varepsilon|\},
\]
with $G(r)=\exp\{-\frac{1}{2}((r-0.5)/0.15)^{2}\}$ and $\kappa_p=0.25\,\kappa_\text{eff}$. During initialization, polarity also supports the double-rate solutions reported in the main paper.

\paragraph{Propeller-centre update (R6).} After every $2\,000$ admitted events for one propeller, their radial Epanechnikov weights $w_e=1-((r_i-1)/\sigma_c)^{2}$ give a~centre measurement. Here $\sigma_c=0.5$; omitting non-positive weights gives support on $r\in(0.5,1.5)$. Writing the admitted event coordinate as $\mathbf z_i=(x_i,y_i)^\top$, we compute from each group
\[
\begin{aligned}
N_\text{eff}&=(\Sigma w)^{2}/\Sigma w^{2},\\
\bar{\mathbf{m}}&=\Sigma_i w_i(x_i,y_i)^\top/\Sigma_i w_i,\\
\mathcal{C}&=\big|\Sigma w\,\mathrm{e}^{\mathrm{i}\theta}\big|/\Sigma w.
\end{aligned}
\]
We omit the update if $N_\text{eff}<50$ or if the angular resultant $\mathcal{C}>0.5$; for two antipodal blade segments, $\mathcal{C}$ measures their imbalance. The predicted propeller centre is $\tilde{\mathbf{p}}=(p_x/p_z,\ p_y/p_z)$ with $\mathbf{p}=\mathbf{H}[\mathbf{c}_k;1]^\top$. The residuals are $\varepsilon_x=\tilde p_x-\bar m_x$ and $\varepsilon_y=\tilde p_y-\bar m_y$, and with $z=1/p_z$,
\[
\begin{aligned}
\mathbf{J}_x&=z\big[p_x-t_x,\ h_{12}c_x-h_{11}c_y,\ 1,\ 0,\ -\tilde p_xc_x,\ -\tilde p_xc_y\big],\\
\mathbf{J}_y&=z\big[p_y-t_y,\ h_{11}c_x+h_{12}c_y,\ 0,\ 1,\ -\tilde p_yc_x,\ -\tilde p_yc_y\big].
\end{aligned}
\]
This measurement uses the variance $R_c=\max\{10^{-6},\ \iota\,\sigma_\text{px}^{2}/N_\text{eff}\}$, where $\sigma_\text{px}=\frac{1}{2}\rho_k\mathrm{e}^{\log s}$ and $\iota=3.0$. We apply the $x$ and $y$ measurements sequentially and recompute the $y$ prediction after the $x$ update.

\paragraph{Blade-edge measurements (R7).}
We tested five measurements that use the shape or polarity of a~blade edge. Each is evaluated for the current event. A~zero weight omits the corresponding measurement, and all five zero-weight controls reproduce their reference rate traces exactly.

\emph{(a) Swept centreline}: not used ($\lambda_\text{swp}=0$). With slope $\varsigma=0.5$,
$\varepsilon=\wrap{\varepsilon^\phi_i-\varsigma(r_i-1)}$,
$\mathbf{J}=\mathbf{J}^{\phi}-\varsigma\,\mathbf{J}^{r}$,
$w=\lambda_\text{swp}\,w_\text{ring}\,\mathrm{e}^{\kappa_\text{eff}(\cos\varepsilon-1)}\,w_\text{Huber}(\varepsilon;c_\phi)$.

\emph{(b) Polarity chord}: not used ($\lambda_\text{chord}=0$). With slope $\varsigma_c=0.5$ and $\mu_p=0.5$, let $\eta=\wrap{\varepsilon^\phi_i-o_{p_i}-p_i\varsigma_c(r_i-\mu_p)}$; then
$\varepsilon=2\sin(\eta/2)$,
$\mathbf{J}=\cos(\eta/2)\big(\mathbf{J}^{\phi}-p_i\varsigma_c\mathbf{J}^{r}\big)$ and
$w=\frac{1}{2}\lambda_\text{chord}\kappa_c G(r_i)^{2}\mathrm{e}^{\kappa_c(\cos\eta-1)}\min\{1,1/|\varepsilon|\}$ with $\kappa_c=0.25\,\kappa_\text{eff}$.

\emph{(c) Rotating radial silhouette}: used with $\lambda_\text{sil}=0.400$. The outer blade envelope is modelled as $b(\varepsilon^\phi)=b_\text{sil}+A\cos\varepsilon^\phi$ with $b_\text{sil}=1.15$, $A=0.25$ and $\tau_s=0.05$; with $d=r_i-b(\varepsilon^\phi_i)$,
\[
\varepsilon=\operatorname{softplus}_{\tau_s}(d),\qquad
w=\lambda_\text{sil}\,w_\text{ring}(r_i),
\]
\[
\mathbf{J}=\sigma(d/\tau_s)\big(\mathbf{J}^{r}+A\sin\varepsilon^\phi_i\,\mathbf{J}^{\phi}\big).
\]
It couples the radial and the angular channel inside one one-sided barrier per event.

\emph{(d) Same-edge transport}: not used ($\lambda_\text{tr}=0$) and defined only for $B=2$. Each propeller stores four previous events, indexed by polarity and by physical blade $\lfloor\phi_k/2\pi\rfloor\bmod 2$. If the corresponding event is less than $10$~ms old,
$\varepsilon=\wrap{\varepsilon^\phi_i-\varepsilon^{\phi,\text{prev}}}$,
$\mathbf{J}=\mathbf{J}^{\phi}-\mathbf{J}^{\phi,\text{prev}}$,
$w=\lambda_\text{tr}\,w_\text{ring}\,w_\text{Huber}(\varepsilon;c_\phi)$.

\emph{(e) Signed crossing}: used with $\lambda_\text{sc}=0.200$. Event polarity must agree with the direction in which the rotating contour crosses the pixel. With convention sign $\chi=+1$, $\tau_x=0.25$ and $d=-\chi\,p_i\sin\varepsilon^\phi_i$,
\[
\varepsilon=\operatorname{softplus}_{\tau_x}(d),\qquad
w=\lambda_\text{sc}\,w_\text{ring}(r_i),
\]
\[
\mathbf{J}=\sigma(d/\tau_x)\,(-\chi\,p_i\cos\varepsilon^\phi_i)\,\mathbf{J}^{\phi}.
\]

We also tested three alternative geometry updates that are not part of the reported method.

\paragraph{Geometry from four propeller centres (R8).} We tested, but do not use, a~similarity measurement in the six-parameter model when all four centre estimates are less than $50$~ms old. Let $\boldsymbol\mu_c$ and $\boldsymbol\mu_m$ denote the point-set means, with centred coordinates $\mathbf{c}'_k=\mathbf{c}_k-\boldsymbol\mu_c$ and $\mathbf{m}'_k=\bar{\mathbf{m}}_k-\boldsymbol\mu_m$. A~closed-form Umeyama fit gives
\[
\hat\psi=\operatorname{atan2}\Big(\sum_k \mathbf{c}'_k\!\times\!\mathbf{m}'_k,\ \sum_k \mathbf{c}'_k\!\cdot\!\mathbf{m}'_k\Big),
\]
\[
\hat s=\frac{\big\|\sum_k(\mathbf{c}'_k\!\cdot\!\mathbf{m}'_k,\ \mathbf{c}'_k\!\times\!\mathbf{m}'_k)\big\|}{\sum_k\|\mathbf{c}'_k\|^{2}},
\]
and $\hat{\mathbf{t}}=\boldsymbol\mu_m-\hat s\,\mathbf{R}(\hat\psi)\boldsymbol\mu_c$. Here $\mathbf{a}\times\mathbf{b}=a_xb_y-a_yb_x$. Four identity-Jacobian rows measure $\log s$, $\psi$, $t_x$ and $t_y$ against $\log\hat s$, $\hat\psi$, $\hat t_x$ and $\hat t_y$; the angular residual is wrapped to $[-\pi,\pi)$. We set $\hat\sigma^2$ to the mean squared fit residual per image coordinate, with a~floor of $0.25$~px$^2$. Translation variance is $\iota\hat\sigma^{2}/4$; log-scale and rotation variance is $\iota\hat\sigma^{2}/\max\{10^{-9},\hat s^{2}\sum_k\|\mathbf{c}'_k\|^{2}\}$, with $\iota=3$.

\paragraph{Similarity constraint (R9).} We tested this constraint only for the full eight-parameter geometry and do not use it in the reported configuration. Every $10$~ms, two identity-Jacobian measurements pull the linear part toward a~similarity,
$\varepsilon_1=h_{11}-h_{22}$ and $\varepsilon_2=h_{12}+h_{21}$, with $R=\sigma_\text{sim}^{2}$.

\paragraph{Windowed geometry update (R10).} A~tested alternative accumulates $\mathbf{A}^\top\mathbf{A}\mathrel{+}=\mathcal{I}\mathbf{J}^\top\mathbf{J}$ and $\mathbf{A}^\top\mathbf{b}\mathrel{+}=\mathcal{I}\,\varepsilon\,\mathbf{J}^\top$, where $\mathcal{I}=w/R_\text{base}$. The reported comparison uses the damped Gauss--Newton update $\Delta\mathbf{q}=-(\mathbf{A}^\top\mathbf{A}+\lambda\diag(\mathbf{A}^\top\mathbf{A})+10^{-12}\mathbf{I})^{-1}\mathbf{A}^\top\mathbf{b}$ with zero forgetting and $\lambda=100$, $100$ and $10$ for the $50$, $100$ and $200$~$\mu$s windows, respectively. We do not use this update because the tested windows reduce the lock percentage.

\begin{table}[t]
\centering
\footnotesize
\setlength{\tabcolsep}{2pt}
\begin{tabular}{@{}p{0.49\columnwidth}p{0.38\columnwidth}c@{}}
\toprule
Measurement & Weight & Used \\
\midrule
R1 helical phase-dynamics (phase) & $\mathcal{I}=\max\{10^{-3},\kappa w_\text{ring}\}$ & \ok \\
R2 geometry helical-phase             & $\lambda_\phi=1$ & \fail \\
R3 radial $(r-1)$, symmetric window   & $\lambda_r=2.000$ & \ok \\
R4 band barriers, gate-following      & $\lambda_\text{band}=4.000$ & \ok \\
R5 polarity                           & $\lambda_\text{pol}=0.500$ & \ok \\
R6 propeller centre, two measurements per $2\,000$ events & none & \ok \\
R7a swept centreline                  & $\lambda_\text{swp}=0$ & \fail \\
R7b polarity chord                    & $\lambda_\text{chord}=0$ & \fail \\
R7c rotating silhouette               & $\lambda_\text{sil}=0.400$ & \ok \\
R7d same-edge transport               & $\lambda_\text{tr}=0$ & \fail \\
R7e signed crossing                   & $\lambda_\text{sc}=0.200$ & \ok \\
R8 Procrustes anchor                  & -- & \fail \\
R9 similarity prior (full 8)          & $\sigma_\text{sim}=0$ & \fail \\
R10 windowed damped-Gauss--Newton step & -- & \fail \\
\bottomrule
\end{tabular}
\caption{Measurements considered in this study. A~check mark indicates that the measurement is used in the reported configuration. The radial, annulus and polarity weights were selected together; rotating silhouette and signed crossing use separately selected weights. The propeller-centre update requires $N_\text{eff}\ge50$ and uses variance inflation $\iota=3.0$.}
\label{tab:suppl_rows}
\end{table}

\section{Configuration used in the experiments}\label{sec:suppl_config}

Table~\ref{tab:suppl_config} gives the scientific parameters used for the fixed-configuration mechanism ablations. The filter state uses blade-passage frequency internally. Every figure, table and reported error converts this quantity to shaft RPM by dividing by the known blade count of two; no accuracy correction is fitted.

\begin{table*}[t]
\centering
\footnotesize
\setlength{\tabcolsep}{3pt}
\begin{tabular}{l l l}
\toprule
Parameter & Value & Role \\
\midrule
\multicolumn{3}{l}{\emph{Measurement weights in the fixed ablation configuration}}\\
$\lambda_r$ (radial) / $\lambda_\text{band}$ / $\lambda_\text{pol}$ & $2.000$ / $4.000$ / $0.500$ & weights of R3 / R4 / R5 \\
$\lambda_\text{sc}$ (signed crossing) / $\lambda_\text{sil}$ (silhouette) & $0.200$ / $0.400$ & weights of R7e / R7c \\
$\lambda_\phi$ (helical phase $\to$ geometry) & $1.0$, measurement omitted & R2 not used \\
radial symmetric window / band gate-following & on / on & measurement support \\
$b_\text{in}$, $b_\text{out}$, $\tau$ (band) & $0.3$, $1.3$, $0.05$ & softplus barriers \\
polarity offsets, gate mean, gate $\sigma$, $\kappa$ scale & $\pm0.25$, $0.5$, $0.15$, $0.25$ & R5 companions \\
signed-crossing $\tau$, polarity sign & $0.25$, $+1$ & R7e companions \\
silhouette outer, amplitude, $\tau$ & $1.15$, $0.25$, $0.05$ & R7c companions \\
\midrule
\multicolumn{3}{l}{\emph{Per-propeller phase block}}\\
$\kappa$ / $q_\text{jerk}$ / $B$ & $22.0$ / $400.0$ / $2$ & von Mises concentration, white-jerk density, blades \\
$R_\phi$ / $\sigma_{\phi_0}$ / $\sigma_{\omega_0}$ / $\sigma_{\alpha_0}$ & $1.0$ / $1.0$ / $10$~rad\,s$^{-1}$ / $100$~rad\,s$^{-2}$ & measurement and initial standard deviations \\
$\sigma_\text{ring}$ & $0.25$ & width of $w_\text{ring}$ \\
rate drift fraction & $0.0025$ & rate random walk, density $(0.0025\,|\omega|)^2$ \\
circular phase-variance clamp & on & ceiling $P_{00}\le\pi^2/3=3.290$ \\
von Mises phase gate on R1 & off & $w_\text{coh}\equiv1$ \\
$\omega$ clamp (helical) / $|\alpha|$ clamp & $[2\,000,\,30\,000]$~RPM / $30\,000$~rad\,s$^{-2}$ & physical clamps \\
\midrule
\multicolumn{3}{l}{\emph{Shared geometry block}}\\
model & reduced 6 & $[\log s,\psi,t_x,t_y,h_{31},h_{32}]$ \\
initial std $(\log s,\psi,\text{affine},t,h_3)$ & $0.05$, $0.05$, $1.0$, $2.0$, $0.005$ & $\mathbf{P}^{q}(t_0)$ \\
process std per $\sqrt{\text{s}}$ & $0.05$, $0.05$, $1.0$, $8.0$, $0.01$ & geometry random walk \\
translation covariance floor & $1.0$~px$^2$ & one-sided; keeps the Kalman gain from vanishing \\
NIS-adaptive process noise & off & \\
per-propeller geometry-measurement budget & off & every eligible measurement is used \\
\midrule
\multicolumn{3}{l}{\emph{Admission and validity}}\\
$r_\text{in}$ / $r_\text{out}$ & $0.2$ / $1.5$ & hard radial gate, propeller-local units \\
image prefilter / pad & on / $1.4$ & four compares before any warp \\
equal radii / no-ring overlap (global, margin) & on / on, $1.0$ & $\rho_k\equiv1$; Voronoi cap on $r_{\text{out},k}$ \\
NIS gate / minimum measurement weight & $16.0$ / $10^{-6}$ & geometry-measurement rejection \\
$|\log s-\log s_0|$ / $|t-t_0|$ / $|h_{3\cdot}|$ & $1.2$ / $2\,000$~px / $0.20$ & sanity bands \\
$|\det\mathbf{H}|$ / $\mathrm{cond}\,\mathbf{H}$ / denominator margin & $\ge10^{-8}$ / $\le10^{8}$ / $\ge0.05$ & geometry checks \\
covariance diagonal floor & $10^{-12}$ & applied to each diagonal entry \\
\midrule
\multicolumn{3}{l}{\emph{Re-acquisition after a~stream gap}}\\
gap threshold / phase-variance reset & $0.020$~s / $3.290$~rad$^2$ & rate kept, phase forgotten \\
$\kappa$ scale / re-acquisition window & $0.15$ / $0.05$~s & widened association gate \\
$\alpha$ reset / geometry freeze & off / off & both reduced lock \\
\midrule
\multicolumn{3}{l}{\emph{Blob ellipse observer and the ring-centring correction}}\\
observer: cells / span / $\tau_\text{fast}$ / $\tau_\text{occ}$ / occ.\ min / $\tau_\text{mom}$ & $28$ / $3.2$ / $0.040$~s / $0.30$~s / $0.85$ / $0.30$~s & estimate-independent blob ellipse \\
servo: gain / baseline $\tau$ / min mass / cap / fold & $5.0$~s$^{-1}$ ($\tau\!=\!0.2$~s) / $10$~s / $500$ / $1.0$ / $0.05$~px & bounded pull toward blob centres \\
\midrule
\multicolumn{3}{l}{\emph{Initialization, sampling and output}}\\
detector pose seed ($\psi$, aspect) & on & seeds $\psi$ and the perspective term \\
$\phi_0$ / seed $\omega_0$ & $0$ / from ground truth at the cut start & initialization used by the mechanism ablations \\
internal blade-pass scale & $2.0$ on both datasets & reported rates divide by $B=2$ \\
event stride / output grid & $1$ / $5$~ms & every event is processed \\
\midrule
\multicolumn{3}{l}{\emph{Fixed constants}}\\
Huber $c_\phi$ / $c_r$ & $0.6$ / $0.3$ & R2, R3, R7a, R7d \\
centroid update interval / $\sigma_c$ / $\iota$ / $N_\text{eff}$ / coverage & $2\,000$~ev.\ / $0.5$ / $3.0$ / $\ge50$ / $\le0.5$ & R6 \\
symmetrization interval & $1\,024$ accepted measurements & $\mathbf{P}^{q}\leftarrow\frac{1}{2}(\mathbf{P}^{q}+\mathbf{P}^{q\top})$ \\
\bottomrule
\end{tabular}
\caption{Fixed configuration used for the mechanism ablations. The primary benchmark instead selects the measurement weights inside each recording-level outer fold. Radii and annulus boundaries use propeller-local units; translations use pixels; $h_{31}$ and $h_{32}$ use inverse propeller-radius units; and angular rates use rad\,s$^{-1}$ unless an~internal limit is stated in blade-pass RPM. Reported rates use shaft RPM.}
\label{tab:suppl_config}
\end{table*}

The rate-process-noise spectral density is $(f|\omega|)^2$, with
$f=0.0025\,\mathrm{s}^{-1/2}$. When predicted phase variance exceeds the
uniform-circular limit, we multiply the first covariance row and column by
$\sqrt{\sigma_{\max}^2/P_{00}}$. This preserves positive semi-definiteness
while setting $P_{00}$ to the limit.

Slow centre drift is handled by an~auxiliary observer outside the event-wise
Kalman measurements. It fits an~ellipse to each propeller's events and compares
the fitted centre with a~running average for that propeller. Bounded translation
and scale corrections then move the shared geometry towards these observed
centres. Their $0.2$~s time constant prevents a~short-lived ellipse fit from
moving the geometry abruptly. Table~\ref{tab:suppl_config} gives the remaining
observer settings.

\section{Geometry validity checks}\label{sec:suppl_validity}

Before accepting a~geometry update, we check that its homography is finite, stable and keeps every modelled propeller in front of the projective horizon. These checks do not establish pose accuracy or prevent drift within their bounds. The initial homography also passes them. Mapping each annotated image centre through $\mathbf{H}_0^{-1}$ gives its body-frame centre $\mathbf{c}_k$, so $\mathbf{H}_0\mathbf{c}_k$ reproduces the annotation.

We accept a~candidate $\mathbf{q}^\text{cand}$, decoded to $\mathbf{H}^\text{cand}$, only if all of the following conditions hold.

\begin{enumerate}[label=(\roman*),leftmargin=1.6em,itemsep=1pt,topsep=2pt]
\item All entries are finite.
\item \textbf{Orientation.} For the full 8-parameter model only, the determinant of the linear part exceeds $10^{-12}$.
\item \textbf{Scale band.} $|\log s-\log s_\text{init}|\le1.2$.
\item \textbf{Translation margin.} $|t_x-t_x^\text{init}|\le2\,000$~px, and likewise for $t_y$. This is a~numerical guard, not a~motion envelope.
\item \textbf{Perspective bound.} $|h_{31}|\le0.20$ and $|h_{32}|\le0.20$.
\item \textbf{Determinant.} $|\det\mathbf{H}^\text{cand}|\ge10^{-8}$.
\item \textbf{Projective denominators} at the four corners of the body bounding box, each $\ge0.05$ (below).
\item \textbf{Condition number} $\sigma_\text{max}/\sigma_\text{min}\le10^{8}$, checked as described below.
\end{enumerate}

Only four corners need checking because the projective denominator is affine in the body point $\mathbf{u}=(u_x,u_y)$:
\[
w(\mathbf{u})=h_{31}u_x+h_{32}u_y+1 .
\]
We require $w(\mathbf{u})\ge0.05$ over every admitted annulus. An~affine function attains its minimum over a~convex set at an~extreme point. Let
\[
\mathcal{U}=\Big[\min_k(c_{kx}-\rho_kr_\text{out}),\ \max_k(c_{kx}+\rho_kr_\text{out})\Big]\times\big[\;\cdot\;\big]_y
\]
be the axis-aligned bounding box of the four annuli, with the $y$ interval defined analogously. It contains every modelled ring point, and its extreme points are its corners, so
\[
\min_{\mathbf{u}\in\mathcal{U}}w(\mathbf{u})=\min_{\mathbf{u}\in\text{corners}(\mathcal{U})}w(\mathbf{u}),
\]
Four comparisons therefore test every modelled ring point, rather than only observed events, and keep them in front of the horizon $w=0$.

To avoid a~singular-value decomposition after every event, we reuse the last checked homography $\mathbf{H}_\text{anchor}$, its extreme singular values and the radius
\[
\varrho=\frac{1}{2}\cdot\frac{\frac{1}{4}\kappa_\text{max}\,\sigma_\text{min}-\sigma_\text{max}}{\frac{1}{4}\kappa_\text{max}+1},
\]
when it is positive. A~candidate within Frobenius distance $\varrho$ needs no new decomposition; otherwise we compute a~Jacobi SVD and replace the anchor. Because such a~perturbation moves each singular value by at most $\varrho$, the radius ensures $(\sigma_\text{max}+\varrho)/(\sigma_\text{min}-\varrho)\le\kappa_\text{max}$. We use a~factor-of-four margin on $\kappa_\text{max}$ and halve $\varrho$ again.

\section{Evaluation details}\label{sec:suppl_scoring}
We use the same scoring grid, reference signals and denominator for all methods and ablations. This section also explains the uncertainty analysis and the event-stream audit.

We measure tracking with the lock percentage. A~propeller is \emph{locked} at time $t$ if $|\widehat{\mathrm{RPM}}(t)/\mathrm{RPM}_\text{GT}(t)-1|\le0.02$. We divide the locked samples by the full nominal $5$~ms grid over all recording--propeller pairs. A~missing sample is therefore unlocked. This matters for \eventpro{}~\cite{chen_count_2026}, which stores a~value only when its checks find a~reliable rate, and for two shortened \helixtrack{}~\cite{spetlik_helixtrack_2026} outputs. The relative tolerance applies equally throughout the evaluated range and excludes estimates at half or twice the true rate.

The TQE reference comes from a~reflective-sensor pulse train and records blade passages; target--follower provides a~direct shaft-rate timeline. We divide the TQE reference and every blade-pass estimate by the known blade count $B=2$. Rates and absolute errors are in shaft RPM; relative errors are dimensionless. This physical conversion fits no scale or timing parameter.

We also audit the target--follower reference independently of the tracker output. The optical-gate pulse polarity alternates in more than $99.94$\,\% of transitions, and timestamp quantization gives a~median rate half-width of $0.6$--$0.9$~RPM. After affine clock alignment, the 95th-percentile timing residual is $3.0$, $3.0$ and $5.3$~ms in the three recordings. In this audit, shifting the reference by $-5$ and $+5$~ms changes lock from $95.87$\,\% to $95.46$\,\% and $95.89$\,\%; shifts of $-10$ and $+10$~ms give $94.74$\,\% and $95.51$\,\%. These checks support the timing and internal consistency of the reference, but they are not an~independent calibration of its absolute rate.

We validated the scorer in two ways. From stored \eventpro{}~\cite{chen_count_2026} samples, it reproduces the recorded mean MAE, MArE and per-recording coverage. It also reproduces the stored lock totals for the recursive-versus-windowed comparison. All methods and ablations use this scorer with the same cuts, grid and ground truth.

\subsection{Missing events and their treatment}\label{sec:suppl_streams}
We define a~damaged stream from the input alone: more than $0.5$~s of complete
sensor silence within any $5$~s interval. This rule is evaluated before examining
any tracker output. Only the $2$~m, $M_\text{ego}=4$ recording meets it.
At $2$~m and $M_\text{ego}=4$, nine empty intervals within $3.5$~s total $664$~ms:
seven last $81$--$85$~ms and two about $40$~ms. At $2$~m and $M_\text{ego}=7$,
four gaps total $333$~ms, below the criterion. The first recording is excluded from selection and the
primary result, then scored as a~separate stress test after both stages are fitted on
the 12 intact moving-camera recordings. The second and the $4$~m, $M_\text{ego}=6$ recording remain
in every outer analysis. On the damaged recording, \ours{}, \helixtrack{}~\cite{spetlik_helixtrack_2026} and \eventpro{}~\cite{chen_count_2026} reach $21.68$\,\%, $15.30$\,\% and $9.11$\,\% lock. \helixtrack{} uses its published ring bounds because this stress case enters no outer fold. Including the recording in a~13-recording aggregate gives $59.69$\,\%, $37.83$\,\% and $21.19$\,\%, so the method ordering is unchanged.

\subsection{Recordings shown in the figures}\label{sec:suppl_visualization}
Both figures use $10$~ms event integrations from recordings named in their captions. Figure~\ref{fig:dataset_progression} uses six timestamps fixed before tracker output was inspected. They begin about $0.5$~s after the fixed evaluation cut of each recording and continue every $0.4$~s. The displayed rings come from the final selected \ours{} configuration. The figures do not define an~evaluation subset.

The first integration windows start at $20.250$~s for the highest-motion $2$~m recording, $14.960$~s for the highest-motion $4$~m recording, and $1.990$~s for the target--follower recording. Each of the six columns advances by $0.4$~s. The first windows start about $0.5$~s after the evaluation cut, rounded upward to a~$10$~ms boundary.

\subsection{Recording-level uncertainty}\label{sec:suppl_uncertainty}
Confidence intervals resample recordings, keeping their four propellers and all timestamps together. We use $100\,000$ paired bootstrap replicates with fixed seeds. On the moving-camera benchmark, \ours{} exceeds the public \helixtrack{} implementation~\cite{spetlik_helixtrack_2026} by $23.18$ percentage points, with a~95\,\% interval of $[9.64,38.77]$ points. Its lead over \eventpro{}~\cite{chen_count_2026} is $40.71$ points, with interval $[27.36,54.44]$. On target--follower, the corresponding differences are $39.66$ points over \helixtrack{}, with interval $[18.01,79.64]$, and $34.78$ points over \eventpro{}, with interval $[25.78,38.96]$. \ours{} has higher lock than both baselines on each of the three target--follower recordings.

\subsection{Lock-threshold sensitivity}\label{sec:suppl_thresholds}
The $2$\,\% threshold is the primary criterion. We also rescore the same trajectories at $1$\,\% and $5$\,\%, without changing configuration selection. Table~\ref{tab:suppl_thresholds} shows that the aggregate ordering remains unchanged on both datasets.

\begin{table}[t]
\centering
\setlength{\tabcolsep}{3pt}
\begin{tabular}{@{}llrrr@{}}
\toprule
Dataset & Method & $1$\,\% & $2$\,\% & $5$\,\% \\
\midrule
Moving camera & \ours{} & \textbf{45.62} & \textbf{62.94} & \textbf{76.79} \\
 & \helixtrack{}~\cite{spetlik_helixtrack_2026} & 26.32 & 39.75 & 48.86 \\
 & \eventpro{}~\cite{chen_count_2026} & 15.22 & 22.22 & 28.84 \\
\midrule
Target--follower & \ours{} & \textbf{32.63} & \textbf{95.87} & \textbf{99.35} \\
 & \helixtrack{}~\cite{spetlik_helixtrack_2026} & 28.53 & 56.21 & 61.01 \\
 & \eventpro{}~\cite{chen_count_2026} & 30.49 & 61.09 & 90.98 \\
\bottomrule
\end{tabular}
\caption{Lock percentage at three relative-error thresholds. No method is retuned between columns.}
\label{tab:suppl_thresholds}
\end{table}

\section{Selection of measurements and weights}\label{sec:suppl_selection}

The moving-camera benchmark repeats both \ours{} configuration stages inside each recording-level outer fold. For each omitted recording, we use the other 11 recordings to compare candidate configurations. We first choose the signed-crossing and rotating-silhouette weights, then keep them fixed while choosing the radial, annulus and polarity weights. Both stages use the fixed $4\,750$ shaft-RPM initialization. The omitted recording contributes to neither stage and is evaluated once. The damaged stream is a~separate stress case and enters no selector. The $4$~m, motion-level-six recording remains an~ordinary outer fold.

The first stage evaluates all $88$ combinations of signed-crossing weights $\{0,.125,.150,.200,.250,.400,.420,.450\}$ and rotating-silhouette weights $\{0,.25,.30,.35,.40,.45,.50,.55,.60,.65,.70\}$. Conditional on the first-stage choice, the second stage evaluates all $48$ combinations of radial weights $\{0,1,2,4\}$, annulus weights $\{0,4,8\}$ and polarity weights $\{0,.25,.50,1.50\}$.

Both prior methods use parameters chosen by grid search on the moving-camera benchmark rather than arbitrary defaults. For \eventpro{}~\cite{chen_count_2026}, we repeat the search inside each outer fold on the same 11 training recordings used for \ours{}; the omitted recording is evaluated only after selection. We give \helixtrack{}~\cite{spetlik_helixtrack_2026} the same outer folds and search the 24 pairs of normalized inner and outer ring bounds reported with that method. The public pair $(0.2,1.5)$ is included as a~control. Selection chooses $(0.1,1.5)$ in seven folds, $(0.2,1.5)$ in four and $(0.1,1.4)$ in one. These held-out evaluations give $39.75$\,\% lock. Applying the public pair to all 12 recordings gives $40.31$\,\%. The nested result is $0.56$ points lower, with a~95\,\% recording-bootstrap interval of $[-1.60,0.02]$ points. The public control reproduces all 48 stored trajectories exactly.

For target--follower, \helixtrack{}~\cite{spetlik_helixtrack_2026} uses its public configuration, while the \ours{} and \eventpro{}~\cite{chen_count_2026} configurations selected on all 12 moving-camera recordings are applied unchanged.

The \eventpro{}~\cite{chen_count_2026} grid contains $243$ combinations: bundle duration $\{10,20,40\}$~ms, polarity-balance tolerance $\{.25,.35,.45\}$, minimum filtered events per bundle $\{32,64,128\}$, maximum bundles per estimate $\{3,6,9\}$ and maximum estimate age $\{125,250,500\}$~ms. Three selected values lay on open boundaries, so we additionally tested the corresponding one-axis extensions: one maximum bundle, $256$ minimum events and $1$~s maximum age. No selected value then lay on an~open boundary.

Let $L_r(c)$ be the locked duration of candidate $c$ on training recording $r$, and let $c_0$ be the reference setting for the current stage. We define $\Delta_r(c)=L_r(c)-L_r(c_0)$. Each stage selects the candidate that maximizes $\sum_r\Delta_r(c)-\max_r\Delta_r(c)$, the smallest total gain remaining after one training recording is removed. Subtracting the largest gain asks how the candidate performs without its most favourable training recording, so one unusually easy recording cannot decide the selection. Ties maximize $\sum_r\Delta_r(c)$ and then use the configuration name. We select the target--follower configuration in the same two stages using all 12 moving-camera recordings, then apply it unchanged to all three target--follower recordings. Table~\ref{tab:suppl_nested_folds} gives every held-out recording and the transferred result.

\begin{table*}[t]
\centering
\setlength{\tabcolsep}{5pt}
\begin{tabular*}{\textwidth}{@{\extracolsep{\fill}}lccrrrr@{}}
\toprule
Evaluation recording & \shortstack{Crossing,\\silhouette} & \shortstack{Radial, annulus,\\polarity} & $T_\text{lock}$ & Lock & MAE & Coverage \\
 & & & [s] & [\%] & [RPM] & [\%] \\
\midrule
$2$~m, $M_\text{ego}=1$ & $(.250,.250)$ & $(2,8,.5)$ & 94.565 & 82.3 & 42.5 & 99.9 \\
$2$~m, $M_\text{ego}=2$ & $(.250,.250)$ & $(2,8,.5)$ & 89.305 & 80.6 & 43.7 & 99.9 \\
$2$~m, $M_\text{ego}=3$ & $(.250,.250)$ & $(0,4,.5)$ & 90.335 & 81.4 & 55.6 & 99.9 \\
$2$~m, $M_\text{ego}=5$ & $(.250,.250)$ & $(2,8,.5)$ & 53.440 & 46.3 & 1565.3 & 99.8 \\
$2$~m, $M_\text{ego}=6$ & $(.200,.450)$ & $(1,4,.25)$ & 43.835 & 39.1 & 7652.8 & 99.9 \\
$2$~m, $M_\text{ego}=7$ & $(.250,.250)$ & $(0,4,.5)$ & 17.935 & 15.7 & 2818.5 & 100.0 \\
\midrule
$4$~m, $M_\text{ego}=1$ & $(.250,.250)$ & $(4,4,.5)$ & 69.175 & 60.3 & 1173.8 & 99.9 \\
$4$~m, $M_\text{ego}=2$ & $(.200,.350)$ & $(2,4,.5)$ & 68.180 & 60.3 & 1170.0 & 99.9 \\
$4$~m, $M_\text{ego}=3$ & $(.250,0)$ & $(2,4,.5)$ & 61.805 & 55.4 & 1391.8 & 99.8 \\
$4$~m, $M_\text{ego}=4$ & $(.250,.250)$ & $(2,8,.5)$ & 94.855 & 82.6 & 81.2 & 99.9 \\
$4$~m, $M_\text{ego}=5$ & $(.250,.250)$ & $(2,8,.5)$ & 87.345 & 80.3 & 151.8 & 99.9 \\
$4$~m, $M_\text{ego}=6$ & $(.250,.250)$ & $(2,8,.5)$ & 83.195 & 72.2 & 450.5 & 99.8 \\
\midrule
Target--follower 1 & $(.250,.250)$ & $(2,8,.5)$ & 206.425 & 92.1 & 66.3 & 100.0 \\
Target--follower 2 & $(.250,.250)$ & $(2,8,.5)$ & 243.685 & 96.5 & 81.1 & 100.0 \\
Target--follower 3 & $(.250,.250)$ & $(2,8,.5)$ & 320.290 & 98.0 & 57.8 & 100.0 \\
\midrule
Moving-camera LORO & & & 853.970 & 62.9 & 1383.1 & 99.9 \\
Target--follower transfer & & & 770.400 & 95.9 & 68.4 & 100.0 \\
\bottomrule
\end{tabular*}
\caption{Two-stage configuration selection with fixed-rate initialization. Each moving-camera row names the configuration selected without that recording. The target--follower rows use one configuration selected on all 12 moving-camera recordings. MAE is in shaft RPM. The damaged $2$~m, $M_\text{ego}=4$ stream is not an~outer fold and is reported separately in Sec.~\ref{sec:suppl_streams}.}
\label{tab:suppl_nested_folds}
\end{table*}

\begin{table}[ht]
\centering
\footnotesize
\setlength{\tabcolsep}{4pt}
\begin{tabular}{ccrr}
\toprule
Crossing & Silhouette & Lock [\%] & $\Delta$ [points] \\
\midrule
off & off & 76.0 & 0.0 \\
on & off & 77.0 & +1.0 \\
off & on & 73.1 & -2.8 \\
on & on & \textbf{82.3} & \textbf{+6.3} \\
\bottomrule
\end{tabular}
\caption{Fixed-configuration factorial for the signed-crossing and rotating-silhouette measurements on the 12 moving-camera and three target--follower recordings. The interaction is $+8.2$ percentage points. These values are not outer-fold scores.}
\label{tab:suppl_edge_factorial}
\end{table}

Table~\ref{tab:suppl_weight_factorial} reports a~separate factorial ablation of the radial, annulus and polarity measurements at fixed signed-crossing and silhouette strengths $(0.200,0.400)$. It explains what the three measurements contribute at this setting, but it is not part of the outer prediction reported above.

\begin{table}[ht]
\centering
\footnotesize
\setlength{\tabcolsep}{3pt}
\begin{tabular}{cccrr}
\toprule
Radial & Band & Polarity & Lock [\%] & $\Delta$ [points] \\
\midrule
off & off & off & 65.5 & -13.7 \\
off & off & on  & 78.7 & -0.6 \\
off & on  & off & 67.2 & -12.0 \\
off & on  & on  & 68.6 & -10.6 \\
on  & off & off & 72.8 & -6.4 \\
on  & off & on  & 79.8 & +0.6 \\
on  & on  & off & 72.2 & -7.0 \\
on  & on  & on  & \textbf{82.3} & \textbf{+3.1} \\
\bottomrule
\end{tabular}
\caption{Fixed-configuration factorial for the radial, annulus and polarity measurements on the 15 intact recordings. Deltas use $(1,4,0.25)$ as the reference. These values are not outer-fold scores.}
\label{tab:suppl_weight_factorial}
\end{table}

\subsection{Propeller-wise four-fold protocol}\label{sec:suppl_propeller_cv}

The moving-camera protocol reported for \helixtrack{}~\cite{spetlik_helixtrack_2026} selects parameters using one propeller identity and evaluates the other three. We reproduce that direction over all 13 moving-camera recordings. Within each fold, we use the training propeller to select the signed-crossing and silhouette strengths, followed by the radial, annulus and polarity weights. The three remaining propellers do not contribute reference errors to either selection stage.

For the second stage, this protocol uses the candidate set $\mathcal G_0\cup\mathcal G_1$, where
\[
\begin{aligned}
\mathcal G_0&=\{0,.25,.50,1,2,4\}\times\{0,1,2,4,8\}\\
&\quad\times\{0,.10,.20,.25,.30,.40,.50\},\\
\mathcal G_1&=\{1,2,4\}\times\{2,4,8\}\times\{.40,.50,.60,.75,1,1.50\},
\end{aligned}
\]
with the coordinates ordered as the radial, annulus and polarity weights.

\begin{table*}[t]
\centering
\setlength{\tabcolsep}{5pt}
\begin{tabular}{ccllrrrr}
\toprule
Training & Test & Crossing, silhouette & Weights $(r,b,p)$ & MAE & MArE & $T_\text{lock}$ & Lock \\
propeller & propellers & & & [RPM] & [\%] & [s] & [\%] \\
\midrule
0 & 1, 2, 3 & $(0.20,0.30)$ & $(0.25,2,0.30)$ & 435.8 & 9.83 & 790.215 & 71.66 \\
1 & 0, 2, 3 & $(0.20,0.30)$ & $(0.50,2,0.25)$ & 667.8 & 14.28 & 744.750 & 67.50 \\
2 & 0, 1, 3 & $(0.125,0.35)$ & $(0.25,1,0.40)$ & 980.6 & 23.99 & 793.770 & 71.94 \\
3 & 0, 1, 2 & $(0.20,0.30)$ & $(0,4,0.40)$ & 633.9 & 13.86 & 759.300 & 68.82 \\
\midrule
\multicolumn{4}{l}{All four folds} & 679.5 & 15.49 & 3088.035 & 69.98 \\
\bottomrule
\end{tabular}
\caption{Complete propeller-wise four-fold result on the 13 moving-camera recordings. Each fold selects both configuration stages using one propeller identity and tests the other three, giving 39 test tracks per fold and 156 track-fold predictions in total. MAE is in shaft RPM, and coverage is $100$\,\% in every fold. Lock normalizes the summed duration over these predictions; each physical propeller track appears in three test folds.}
\label{tab:suppl_propeller_cv}
\end{table*}

This protocol holds out propeller identity, not recordings. Every fold retains the same camera motions, and the four propellers share one geometry state. It therefore measures transfer across propeller identity within the known recordings, whereas the primary protocol tests transfer to an~omitted recording.

We report it because it matches the protocol used with \helixtrack{}~\cite{spetlik_helixtrack_2026}. The damaged $2$~m, $M_\text{ego}=4$ stream remains included to match the 13-recording benchmark, and the $4$~m, $M_\text{ego}=6$ recording is included. The aggregate reaches $69.98$\,\% lock, with $100$\,\% coverage, $679.5$ shaft-RPM MAE and $15.49$\,\% MArE.

\begin{table*}[t]
\centering
\setlength{\tabcolsep}{5pt}
\begin{tabular*}{\textwidth}{@{\extracolsep{\fill}}llrrr@{}}
\toprule
Dataset & Initialization & $T_\text{lock}$ [s] & Lock [\%] & MAE [RPM] \\
\midrule
Moving camera & reference rate & 857.9 & 63.2 & 834.5 \\
              & fixed $4\,750$~RPM & 854.0 & 62.9 & 1383.1 \\
\midrule
Target--follower & reference rate & 770.44 & 95.874 & 64.2 \\
                 & fixed $9\,500$~RPM & 770.40 & 95.869 & 68.4 \\
\bottomrule
\end{tabular*}
\caption{\textbf{Rate initialization for \ours{}.} For each moving-camera recording, both rows use the configuration selected on the other 11 recordings and differ only in their initial rate. The target--follower rows use the configuration selected on all 12 moving-camera recordings. The fixed-rate rows are the reported setting; the reference-rate rows are a~diagnostic comparison, not a~main result. The fixed values are $4\,750$ shaft RPM for moving camera and $9\,500$ shaft RPM for target--follower.}\label{tab:ablation}
\end{table*}

On the moving-camera recordings, fixed rather than reference-rate initialization changes lock by $-0.29$ points, with a~95\,\% interval of $[-7.96,7.98]$ points. On target--follower, the difference is $-0.005$ points, with an~interval of $[-0.099,0.083]$ points. These results do not show that access to the first reference rate improves lock.

\section{Comparison protocols}\label{sec:suppl_arms}

\subsection{Baseline methods}\label{sec:suppl_eventpro}
Our \eventpro{}~\cite{chen_count_2026} reimplementation follows its published event-count and polarity maps, rate and direction search, and five exposed constants. We select these constants by grid search inside the same leave-one-recording-out training folds used for \ours{}, with the published setting included as a~control. When the best value lies at an~open search boundary, we extend that direction before scoring the omitted recording. The selected setting reaches $22.22$\,\% lock on the moving-camera benchmark and transfers unchanged to $61.09$\,\% on target--follower. When EventPro's internal checks do not find a~reliable rate, it reports no value; the common grid counts that interval as unlocked. The EventPro paper provides neither public code nor evaluation data, so we did not compare our implementation with outputs from its authors. We therefore identify these numbers as results of our reimplementation.

\paragraph{\helixtrack{}~\cite{spetlik_helixtrack_2026}.}
We measure the public \helixtrack{} implementation once per propeller, using the same cuts, $5$~ms grid, ring annotations and fixed initial rate as \ours{}. On the moving-camera benchmark, each outer fold selects the normalized ring bounds from the published grid without the recording it scores. On target--follower, we retain the published configuration. We initialize the two-blade internal state at $9\,500$ blade-pass RPM on TQE and $19\,000$ blade-pass RPM on target--follower, giving starts of $4\,750$ and $9\,500$ shaft RPM. We divide every output by $B=2$ before reporting it. These measurements are our replay of the public implementation, not results certified by its authors.

\paragraph{Short \helixtrack{}~\cite{spetlik_helixtrack_2026} outputs.}\label{sec:suppl_helixtrack_prefix}
Two of the 64 fixed-rate runs end before the canonical scoring grid; repeated runs end at the same samples. We retain these outputs and count their missing suffixes as unlocked, following the common scoring rule. To bound their effect in the direction most favourable to \helixtrack{}, we also count every missing suffix sample as locked. This gives upper bounds of $40.54$\,\% on the 12 moving-camera recordings and $58.22$\,\% on the three target--follower recordings, compared with $62.94$\,\% and $95.87$\,\% for \ours{}. The ordering is therefore independent of how these suffixes are treated.

Each ablation below has its own matched control. Within a~paragraph, the control and intervention use the same evaluation and differ only in the named choice. Control scores differ slightly across paragraphs, so only the change within one paragraph answers its ablation question.

\begin{table*}[t]
\centering
\setlength{\tabcolsep}{4pt}
\renewcommand{\arraystretch}{1.08}
\begin{tabular}{@{}>{\raggedright\arraybackslash}p{0.48\linewidth}@{\hspace{0.038333\linewidth}}w{r}{0.135\linewidth}@{\hspace{0.038333\linewidth}}w{r}{0.135\linewidth}@{\hspace{0.038333\linewidth}}w{r}{0.135\linewidth}@{}}
\toprule
Setting & $T_\text{lock}$ & Lock \% & MAE \\
\midrule
\multicolumn{4}{@{}l}{\emph{Propellers contributing to shared geometry}} \\
one & 45.5 & 3.4 & 10070 \\
two & 240.8 & 17.7 & 6954 \\
three & 680.5 & 50.2 & 1812 \\
four & \best{854.0} & \best{62.9} & \best{1383} \\
\addlinespace[2pt]
\midrule
\multicolumn{4}{@{}l}{\emph{Kalman and windowed Gauss--Newton geometry}} \\
per-event Kalman control & \best{855.1} & \best{63.0} & \best{1904} \\
selected Gauss--Newton window & 541.0 & 39.9 & 2193 \\
\addlinespace[2pt]
\midrule
\multicolumn{4}{@{}l}{\emph{Stored uncertainty}} \\
matched control, correlations cleared & \best{859.5} & \best{63.3} & \best{875} \\
full covariance & 830.8 & 61.2 & 2223 \\
\addlinespace[2pt]
\midrule
\multicolumn{4}{@{}l}{\emph{Homography model}} \\
full eight-parameter & 801.5 & 59.1 & \best{1088} \\
restricted six-parameter & \best{854.0} & \best{62.9} & 1383 \\
\bottomrule
\end{tabular}
\caption{\textbf{Complete results for the controlled comparisons} summarized in Table~\ref{tab:core_ablations}. Each block has its own control, and only the named choice changes within that block. $T_\text{lock}$ is in propeller-seconds, and MAE is in shaft RPM.}\label{tab:core_ablations_full}
\end{table*}

\paragraph{One-propeller ablation.}
For each tested recording, we use the configuration selected on the other 11 recordings. All four rate states remain active, while the comparison changes which propellers contribute geometry measurements. Results from subsets of equal size are averaged so that every propeller contributes equally. With one, two, three and four contributors, lock is $3.4$\,\%, $17.7$\,\%, $50.2$\,\% and $62.9$\,\%. Four rather than one raises lock by $59.6$ points, with a~95\,\% recording-bootstrap interval of $[47.8,70.1]$ points and higher lock on all 12 recordings. A~single annulus weakly constrains in-plane rotation $\psi$ because its phase residual absorbs body rotation into that propeller's phase. This comparison changes both the spatial support and the amount of geometry evidence. It measures their combined effect and is not a~separately optimized single-propeller tracker.

\paragraph{Slow ellipse observer.}
The observer maintains rotor-local event activity, occupancy and decayed second moments. It fits an~ellipse to the occupied cells and compares its centre with a~slow reference for that propeller. These differences provide bounded scale and translation corrections to the shared geometry. Its effect depends on the measurement weights. Repeating the complete moving-camera selection without it improves held-out lock by $4.9$ percentage points, but the selected configuration loses $30.3$ points after transfer to target--follower relative to the reported method.

To separate the observer from the weights, we crossed its enabled and disabled states with the two selected weight configurations on target--follower. With the reported weights, changing the observer state changes lock by less than $0.01$ percentage points. With the weights selected while the observer was disabled, enabling it adds $29.9$ points. This difference is concentrated in one recording and does not establish a~uniform benefit. The reported observer-enabled method was fixed before this diagnostic.

\paragraph{Six- and eight-parameter homographies.}
The reported model restricts the linear part of the homography to a~scale and rotation. The full model also estimates independent affine scale and shear; all other settings remain fixed. The restricted model gives $3.9$ percentage points more lock, with a~95\,\% interval of $[-6.5,13.4]$ points and wins and losses on $8/4$ recordings. The interval does not establish which model is better, so we retain the smaller six-parameter model.

\paragraph{Fixed-rate initialization.}
The main comparison selects and evaluates \ours{} with a~fixed start of $4\,750$ shaft RPM on TQE and $9\,500$ shaft RPM on target--follower. The evaluated recording's rate therefore affects neither selection nor initialization. Under the same fixed-rate condition, the public \helixtrack{}~\cite{spetlik_helixtrack_2026} implementation gives $39.8$\,\% lock on TQE and $56.2$\,\% on target--follower.

In a~separate fixed-configuration sweep, we replace the ground-truth initial rate with $3\,000$, $4\,750$ or $6\,500$ shaft RPM on TQE and $6\,000$, $9\,500$ or $13\,000$ shaft RPM on target--follower. Pose still starts from the rate-free ring annotation, with $\phi_0=0$. A~track acquires at its first one-second interval with at least 95\,\% locked samples; otherwise it is unrecovered. Before scoring this sweep, we reproduce all 64 ground-truth-initialized trajectories exactly.

On the moving-camera benchmark, ground-truth initialization and the three fixed
rates recover $51$, $50$, $48$ and $49$ of $52$ tracks, respectively. Their
	median acquisition times are $0.310$, $0.397$, $0.297$ and $0.650$~s. On the
target--follower dataset, all $12$ tracks recover; the corresponding median
	times are $0.005$, $0.007$, $0.003$ and $0.005$~s.

\paragraph{Recursive versus windowed geometry.}
The recursive and windowed versions use the same admission rules, state, phase prediction, measurement definitions, initialization and output grid. Geometry changes from sequential Kalman updates to damped Gauss--Newton normal equations accumulated within a~window (Sec.~\ref{sec:suppl_rows}). The ellipse observer remains active in both versions. Both the solver and update schedule change. The two geometry estimates determine the map used for later event admission and assignment. Fixed windows of $50$, $100$ and $200$~$\mu$s give $22.0$\,\%, $45.2$\,\% and $39.7$\,\% lock. For each tested recording, the other 11 select among these windows: $100$~$\mu$s is selected in eight folds and $200$~$\mu$s in four. Per-event Kalman geometry updates improve held-out lock by $23.1$ points over the selected windows. The 95\,\% interval is $[3.7,43.5]$ points, with wins and losses on $9/3$ recordings.

We also screened fixed windows of $2$, $3.2$, $5$ and $10$~ms on three recordings chosen before examining these results. These durations span the approximate half-revolution timescale over the evaluated rate range. Each retained only $38$--$40$\,\% of the recursive version's locked time and failed the predefined $95$\,\% retention test on every recording, so none was promoted to the full comparison.

\paragraph{Block versus full covariance.}
The original block-specific Jacobians produce zero phase--geometry covariance even in a~dense matrix. We therefore derive joint Jacobians for polarity, signed crossing and rotating silhouette. Both dense versions use them; one clears the cross-covariance blocks after each update and the other retains them. Recorded values confirm this difference. Retaining the correlations changes lock by $-2.1$ points, with a~95\,\% interval of $[-13.8,6.8]$ points and wins and losses on $7/5$ recordings. The interval does not establish a~benefit from the larger representation, so the reported filter retains the smaller covariance blocks. The split-measurement control changes both factorization and covariance representation, so we report it separately.

\paragraph{Measurements from the same event.}\label{sec:suppl_event_dependence}
The reported filter applies every active scalar measurement produced by an~event. These measurements use the same event coordinates and predicted state. As a~control, a~deterministic hash of the event time, position, polarity and assigned propeller selects one active measurement family without using its error. We multiply that family's information by the number of active families, preserving each family's expected contribution without reusing an~event. Both versions run in the same executable and reproduce their controls exactly. Applying all measurements improves lock by $1.88$ points over the one-measurement control. The 95\,\% recording-bootstrap interval is $[-3.33,6.38]$ points, with wins, ties and losses on $8/1/3$ recordings. Normalized innovations change substantially between the two versions. We therefore use the sequential measurements for tracking, but do not treat the stored covariance as calibrated.

\paragraph{Runtime protocol.}\label{sec:suppl_runtime}
We measure \ours{} and the public \helixtrack{} implementation~\cite{spetlik_helixtrack_2026} with the same end-to-end wall-clock boundary. Both process every event with one CPU and one numerical-library thread. Each recording stays on one node, and every full-session run is repeated twice in balanced order. \ours{} returns four rates in one run; \helixtrack{} returns one, so we sum its four sequential channel times before computing the four-rate real-time factor. We never add four single-channel real-time factors.

Table~\ref{tab:suppl_runtime_all} reports all 15 recordings. The geometric-mean four-rate factors are $0.0306$ for \ours{} and $0.0189$ for \helixtrack{}~\cite{spetlik_helixtrack_2026} on the moving-camera benchmark, and $0.0478$ and $0.0593$ on target--follower. Thus \ours{} is $1.62\times$ faster on the moving-camera recordings, while \helixtrack{} is $1.24\times$ faster on target--follower. Neither implementation reaches real time. These measurements include configuration parsing, file input and output, SDK decoding and estimation. At $2$~m, the median input rate rises from $39.6$ to $88.1$~Mev/s with camera motion. For \ours{}, wall time nevertheless follows the admitted-event count more closely than the total input rate because the phase and geometry updates after admission dominate the profile.

\begin{table}[t]
\centering
\setlength{\tabcolsep}{3pt}
\begin{tabular}{@{}lrrr@{}}
\toprule
Recording & \ours{} & \helixtrack{}~\cite{spetlik_helixtrack_2026} & Speed ratio \\
\midrule
$2$~m, $M_\text{ego}=1$ & .0210 & .0237 & .88 \\
$2$~m, $M_\text{ego}=2$ & .0183 & .0196 & .93 \\
$2$~m, $M_\text{ego}=3$ & .0195 & .0176 & 1.11 \\
$2$~m, $M_\text{ego}=5$ & .0241 & .0122 & 1.98 \\
$2$~m, $M_\text{ego}=6$ & .0215 & .0138 & 1.56 \\
$2$~m, $M_\text{ego}=7$ & .0268 & .0107 & 2.50 \\
$4$~m, $M_\text{ego}=1$ & .0471 & .0355 & 1.33 \\
$4$~m, $M_\text{ego}=2$ & .0462 & .0355 & 1.30 \\
$4$~m, $M_\text{ego}=3$ & .0477 & .0221 & 2.16 \\
$4$~m, $M_\text{ego}=4$ & .0348 & .0257 & 1.35 \\
$4$~m, $M_\text{ego}=5$ & .0385 & .0142 & 2.71 \\
$4$~m, $M_\text{ego}=6$ & .0461 & .0136 & 3.38 \\
Target--follower 1 & .0253 & .0412 & .61 \\
Target--follower 2 & .0621 & .0641 & .97 \\
Target--follower 3 & .0695 & .0791 & .88 \\
\bottomrule
\end{tabular}
\caption{End-to-end real-time factor for four rates. Higher is faster. Speed ratio is \helixtrack{}~\cite{spetlik_helixtrack_2026} wall time divided by \ours{} wall time, so values above one favour \ours{}. Each comparison uses one node, one CPU thread and two balanced repeats.}
\label{tab:suppl_runtime_all}
\end{table}

\subsection{Sensitivity to ring calibration}\label{sec:suppl_annotations}
We test sensitivity to supplied ring calibration with the selected measurement settings fixed. Across all 12 intact TQE and three target--follower recordings, independent centre offsets equal $1$, $2$, $5$ or $10$\,\% of each recording's mean annotated radius. These magnitudes span $0.3$ to $4.5$ pixels across the two datasets. Four fixed random direction realizations per recording use the same directions at all four magnitudes. We also multiply all supplied radii by $0.80$, $0.90$, $0.95$, $1.05$, $1.10$ or $1.20$. Each condition retains the observer, initial rate and complete scoring interval.

The supplied centres and radii determine the fixed rotor-plane layout through the initial homography. Both perturbations change the fixed layout and initial image geometry. We change the supplied pixel radius while holding the radial measurement weight fixed. The nominal ring annotations define the reference inputs, and these controlled changes do not estimate a~human annotation-error distribution. We do not reselect settings for each altered calibration, so this test measures the reported configuration's response to these supplied inputs. We average direction realizations within each recording before paired recording resampling.

Table~\ref{tab:annotation_perturbations} reports every magnitude. TQE lock ranges from $55.4$\,\% to $66.8$\,\%, compared with $62.9$\,\% nominally. Target--follower is more sensitive to radius increases. Only three recordings enter the resampling.

Target--follower radius sensitivity is concentrated in one of its three recordings. On that recording, increasing all supplied radii by $5$\,\% reduces lock from $96.5$\,\% to $0.4$\,\%, despite $100$\,\% coverage. Centre sensitivity is also non-monotonic: one realization gives $0.8$\,\% lock at $5$\,\% displacement and $96.8$\,\% at $10$\,\%. These responses do not establish a~monotonic tolerance bound for the supplied calibration.

\begin{table*}[t]
\centering
\begin{tabular}{lrrrr}
\toprule
Annotation change & TQE lock [\%] & Change [points] & Target lock [\%] & Change [points] \\
\midrule
Nominal & 62.9 & --- & 95.9 & --- \\
Centre displacement 1\,\% & 66.4 & +3.4 $[-1.0,8.3]$ & 95.9 & +0.0 $[-0.0,0.1]$ \\
Centre displacement 2\,\% & 63.5 & +0.6 $[-4.8,6.5]$ & 95.9 & -0.0 $[-0.1,0.1]$ \\
Centre displacement 5\,\% & 61.0 & -1.9 $[-11.7,5.9]$ & 88.4 & -7.5 $[-23.9,0.0]$ \\
Centre displacement 10\,\% & 60.9 & -2.0 $[-7.4,3.0]$ & 95.9 & -0.0 $[-0.1,0.1]$ \\
\midrule
Radius multiplier $0.80$ & 55.4 & -7.6 $[-17.2,0.4]$ & 94.9 & -1.0 $[-1.9,-0.4]$ \\
Radius multiplier $0.90$ & 63.9 & +0.9 $[-3.5,6.4]$ & 95.8 & -0.1 $[-0.1,-0.0]$ \\
Radius multiplier $0.95$ & 64.0 & +1.0 $[-6.3,8.1]$ & 95.9 & +0.0 $[-0.0,0.1]$ \\
Radius multiplier $1.05$ & 66.8 & +3.8 $[-0.2,8.7]$ & 65.7 & -30.2 $[-96.0,0.0]$ \\
Radius multiplier $1.10$ & 56.8 & -6.1 $[-23.4,8.5]$ & 65.6 & -30.3 $[-96.3,-0.1]$ \\
Radius multiplier $1.20$ & 61.8 & -1.1 $[-11.6,10.3]$ & 65.3 & -30.5 $[-96.4,-0.2]$ \\
\bottomrule
\end{tabular}

\caption{Sensitivity to supplied ring calibration on 12 TQE and three target--follower recordings, with selected measurement settings fixed. Lock scores all four rates at $2$\,\% relative tolerance on the complete $5$~ms grid; missing estimates are unlocked. Centre displacements are fractions of the nominal mean ring radius. Scores average four direction realizations within each recording. Radius multipliers apply to all four supplied rings. Changes are percentage points from the nominal row; brackets give 95\,\% intervals from $100\,000$ paired recording resamples, with time denominators recomputed in each draw. Both perturbations change the assumed fixed layout.}
\label{tab:annotation_perturbations}
\end{table*}

\subsection{Contributor comparison with fixed budgets}\label{sec:suppl_budgets}
The contributor ablation above changes both spatial support and measurement quantity. We therefore add a~diagnostic with the same total geometry-event opportunity budget for one and four contributors. All four phase states remain active and use every admitted event. The geometry gate follows the phase update and precedes centre accumulation, so an~event excluded by the budget supplies no geometry measurements or centre-accumulation input. Both ellipse-observer switches are off because their corrections would otherwise bypass this budget. We retain the weights and initial rates.

We test caps of $2\,000$, $10\,000$, $100\,000$ and $1\,000\,000$ opportunities per second. Successive fixed event-time slots belong to the allowed contributors in round-robin order. The first admitted event from the designated contributor supplies its geometry measurements. An~empty slot expires without carryover. Slots start at the benchmark cut. If an~event precedes the latest time seen by this gate, we discard its geometry input and retain its phase update.

Slot assignment uses neither measurement residuals nor reference rates. We evaluate every single-contributor mask and the four-contributor mask on all 12 intact TQE recordings. Single-contributor scores are averaged within each recording before comparison and recording resampling.

This rule equalizes scheduled opportunities. Spatial admission and measurement rejection remain state-dependent, so filled slots and accepted scalar measurements differ between contributor conditions. We report realized counts with the lock results. Equal event counts alone would not imply equal geometric information. The unbudgeted control uses the same executable and settings with the observer disabled.

The $2\,000$/s and $10\,000$/s caps leave both contributor conditions near a~tracking floor. We fixed the two higher caps after observing this floor and retain all four comparisons in Table~\ref{tab:contributor_budgets}. Every condition remains at or below $2.9$\,\% aggregate lock, versus $62.3$\,\% without a~cap. Although the lowest cap gives a~positive interval for Four minus Single, this small difference does not restore useful tracking. The maximum selected-count gaps against the mean single count range from $1.3$\,\% to $87.1$\,\%. The experiment therefore establishes neither a~practical contributor advantage under these caps nor a~benefit caused by spatial support alone. Table~\ref{tab:contributor_budget_counts} gives each contributor's realized input.

\begin{table*}[t]
\centering
\begin{tabular}{rrrrrrrr}
\toprule
Slots/s & \multicolumn{2}{c}{Lock [\%]} & Change [points] & \multicolumn{2}{c}{Filled slots [\%]} & \multicolumn{2}{c}{Max. count gap [\%]} \\
 & Single & Four & & Single & Four & Mean & Individual \\
\midrule
2000 & 1.2 & 2.9 & +1.7 $[0.3,3.5]$ & 99.8 & 99.6 & 1.3 & 1.6 \\
10000 & 1.1 & 1.4 & +0.3 $[-0.3,1.3]$ & 93.6 & 83.7 & 34.7 & 39.1 \\
100000 & 1.3 & 1.5 & +0.2 $[-0.4,1.1]$ & 60.8 & 40.5 & 74.5 & 76.5 \\
1000000 & 1.7 & 1.6 & -0.1 $[-1.0,1.1]$ & 36.1 & 23.7 & 87.1 & 91.5 \\
\bottomrule
\end{tabular}

\caption{Contributor comparison under four equal geometry-event caps on all 12 TQE recordings, with the ellipse observer disabled. All four rate states enter the $2$\,\% lock score; missing estimates are unlocked. Single averages the four single-contributor scores within each recording. Change is Four minus Single in percentage points, with 95\,\% intervals from $100\,000$ paired recording resamples. Filled slots are selected events divided by scheduled opportunities. Count gaps are absolute selected-count differences divided by the larger count. Mean gives the maximum gap against the mean single count over recordings; Individual also maximizes over single masks. Equal caps do not imply equal realized measurements.}
\label{tab:contributor_budgets}
\end{table*}

\begin{table*}[t]
\centering
\begin{tabular}{rrrrr}
\toprule
Slots/s & Contributor(s) & Lock [\%] & Selected events [$10^3$] & Accepted rows [$10^3$] \\
\midrule
2000 & 0 & 1.3 & 677.1 & 2166.8 \\
2000 & 1 & 1.3 & 677.1 & 2000.0 \\
2000 & 2 & 1.0 & 677.0 & 2115.2 \\
2000 & 3 & 1.2 & 676.5 & 2176.8 \\
2000 & all four & 2.9 & 675.4 & 2238.1 \\
10000 & 0 & 0.9 & 3089.1 & 9359.1 \\
10000 & 1 & 1.2 & 3239.9 & 8353.3 \\
10000 & 2 & 1.0 & 3232.7 & 8890.7 \\
10000 & 3 & 1.4 & 3137.5 & 8919.0 \\
10000 & all four & 1.4 & 2837.7 & 9389.6 \\
100000 & 0 & 1.2 & 21813.1 & 54544.1 \\
100000 & 1 & 1.3 & 19035.5 & 33486.0 \\
100000 & 2 & 1.4 & 22440.9 & 49878.7 \\
100000 & 3 & 1.5 & 19241.3 & 46437.9 \\
100000 & all four & 1.5 & 13742.7 & 45759.2 \\
1000000 & 0 & 1.5 & 140189.4 & 357058.6 \\
1000000 & 1 & 2.2 & 71806.9 & 131139.8 \\
1000000 & 2 & 1.5 & 139627.5 & 280961.7 \\
1000000 & 3 & 1.8 & 137732.2 & 249809.3 \\
1000000 & all four & 1.6 & 80442.2 & 267272.1 \\
\bottomrule
\end{tabular}

\caption{Realized geometry input for each contributor condition in Table~\ref{tab:contributor_budgets}. Single propellers use annotation indices $0$--$3$. Lock scores all four rates, including those that supply no geometry input. Selected events and accepted scalar geometry rows are totals over the 12 TQE recordings, in thousands. One selected event produces one or more scalar rows, and measurement rejection remains active. These counts exclude phase updates, which use every admitted event.}
\label{tab:contributor_budget_counts}
\end{table*}

\section{Additional negative results}\label{sec:suppl_negative}

\paragraph{Iterated relinearization.}
Relinearizing the radial measurement $2$, $3$ or $5$ times gives bit-identical rate traces. We therefore retain one radial linearization per event. This test concerns the radial measurement and does not compare iterations of the windowed solver.

\paragraph{\eventpro{}~\cite{chen_count_2026} lower-MAE cases.}
Our selected \eventpro{} reimplementation has lower shaft-rate MAE than \ours{} at $2$~m, $M_\text{ego}=6$ and $7$, and at $4$~m, $M_\text{ego}=1$, $2$ and $3$. \ours{} has higher lock on each of these recordings. Table~\ref{tab:benchmark_ego} reports both measures and coverage, preserving the cases that favour the baseline on mean error.

\end{document}